\documentclass{article}

\usepackage{PRIMEarxiv}

\usepackage[utf8]{inputenc} 
\usepackage[T1]{fontenc}    
\usepackage{float}
\usepackage{hyperref}       
\usepackage{url}            
\usepackage{booktabs}       
\usepackage{amsfonts}       
\usepackage{nicefrac}       
\usepackage{microtype}      
\usepackage{lipsum}
\usepackage{fancyhdr}       
\usepackage{graphicx}       
\usepackage{amsmath}
\graphicspath{{media/}}     
\usepackage{multirow} 
\usepackage{mathtools}
\usepackage{svg}
\usepackage{xcolor}
\usepackage{makecell}
\usepackage{geometry}

\usepackage{adjustbox}
\usepackage{threeparttable}
\usepackage{subcaption}

\usepackage{fancyhdr}

\title{\textbf{BenchCloudVision}: A Benchmark Analysis of Deep Learning Approaches for Cloud Detection and Segmentation in Remote Sensing Imagery}

\author{
  Fabio Loddo*\\
  \texttt{fabio.loddo@supsi.ch} \\
  \And
  Piga Dario \\
  \texttt{dario.piga@supsi.ch} \\
  \And
  Umberto Michelucci \\
  \texttt{umberto.michelucci@hslu.ch} \\
  \And
  Safouane EL GHAZOUALI*\\
  \texttt{safouane.elghazouali@toelt.ai} \\
}

\begin{document}
\maketitle

\begin{abstract}
    Satellites equipped with optical sensors capture high-resolution imagery, providing valuable insights into various environmental phenomena. In recent years, there has been a surge of research focused on addressing  some challenges in remote sensing, ranging from water detection in diverse landscapes to the segmentation of mountainous and terrains. Ongoing investigations goals to enhance the precision and efficiency of satellite imagery analysis. Especially, there is a growing emphasis on developing methodologies for accurate water body detection, snow and clouds, important for environmental monitoring, resource management, and disaster response. Within this context, this paper focus on the cloud segmentation from remote sensing imagery. Accurate remote sensing data analysis can be challenging due to the presence of clouds in optical sensor-based applications. The quality of resulting products such as applications and research is directly impacted by cloud detection, which plays a key role in the remote sensing data processing pipeline. This paper examines seven cutting-edge semantic segmentation and detection algorithms applied to clouds identification, conducting a benchmark analysis  to evaluate their architectural approaches and identify the most performing ones. To increase the model's adaptability, critical elements including the type of imagery and the amount of spectral bands used during training are analyzed. Additionally, this research tries to produce machine learning algorithms that can perform cloud segmentation using only a few spectral bands, including RGB and RGBN-IR combinations. The model's flexibility for a variety of applications and user scenarios is assessed by using imagery from Sentinel-2 and Landsat-8 as datasets. This benchmark can be reproduced using the material from this github link: \url{https://github.com/toelt-llc/cloud\_segmentation\_comparative}.
\end{abstract}


\section{Datasets}\label{datasets}
\begin{figure}[h]
    \begin{center}
        \includegraphics[width=\textwidth]{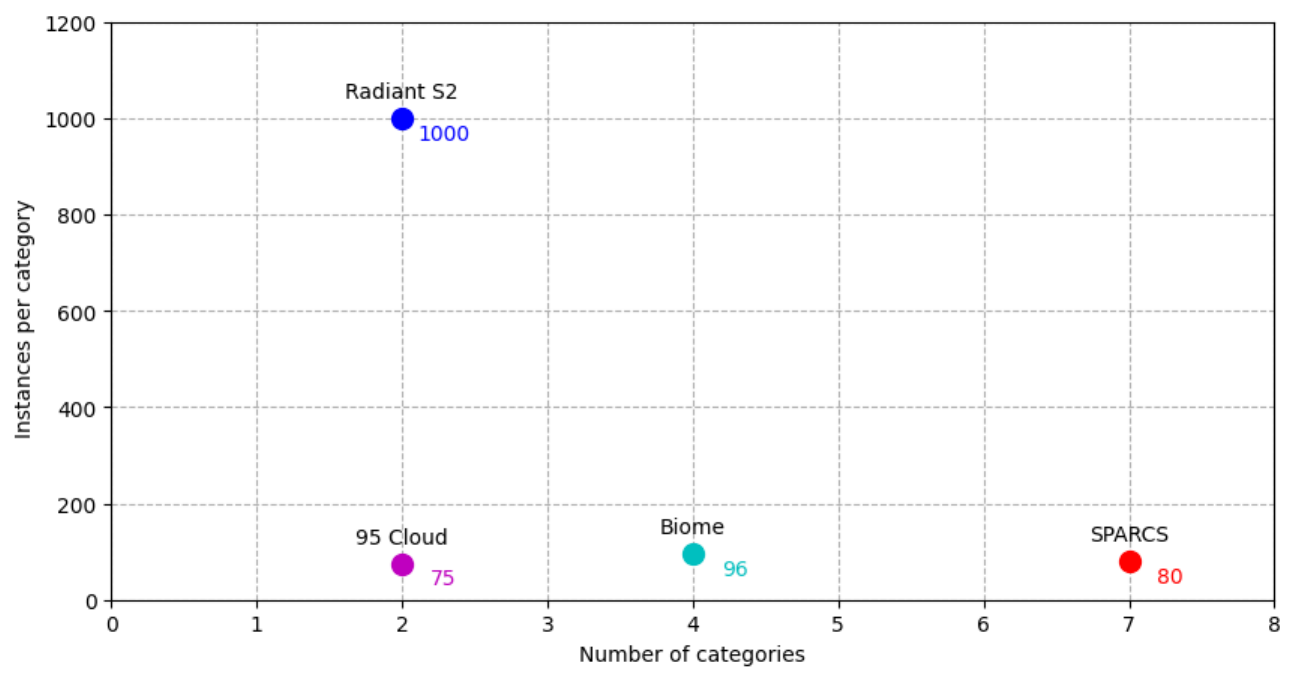}
    \end{center}
    \caption{Display of the number of instances (images) in each set versus the respective number of categories.}
    \label{fig:datasets}
\end{figure}

The current study involves a thorough benchmark analysis, evaluating modern deep learning models for cloud detection in remote sensing imagery. The principal objective encompasses the provision of a meticulous and relative evaluation of these models, offering elucidations regarding their proficiencies, deficiencies, and potential deployment utility. This endeavour is facilitated through the utilization of an array of publicly available datasets, each offering singular challenges and scenarios for cloud detection. The datasets (Fig.\ref{fig:datasets}) used in these papers are:

\begin{itemize}
    \item \textbf{The Biome Dataset:} a meritorious resource introduced by Foga et al. \cite{foga_cloud_2017}, the Biome dataset spans a spectrum of remote sensing imagery, proffering a veracious portrayal of cloud cover across diverse geographic locales and climates.
    
    \item \textbf{The SPARCS Dataset:} also propounded by Foga et al. \cite{foga_cloud_2017}, the SPARCS dataset augments the ambit of evaluation, incorporating disparate conditions and data provenance, thus enhancing the benchmark's granularity.
    
    \item \textbf{The 95-Cloud Dataset:} as introduced by Mohajerani et al. \cite{mohajerani_cloud_2018}, the 95-Cloud dataset broadens the evaluative spectrum, integrating additional challenges and cloud cover variations.
    
    \item \textbf{The Radiant Earth Foundation Sentinel-2 Dataset:} deriving from the extensive gamut of Sentinel-2 satellite imagery, this dataset \cite{mlhub_sentinel_2} affords an abundance of high-resolution data, facilitating the assessment of deep learning models within the context of cloud detection on a broader scale.
\end{itemize}

As reported by Jeppesen et al. in their work \cite{jeppesen_cloud_2019}, an analysis of the dataset distributions reveals significant dissimilarities between the Biome and SPARCS datasets. Notably, the SPARCS dataset exhibits a higher prevalence of clear scenes compared to the Biome dataset. Furthermore, the distribution patterns suggest that the Biome dataset places a pronounced emphasis on three specific categories: clear, mid-cloudy, and cloudy scenes. This emphasis may not be entirely representative of Landsat 8 data as a whole.

One noteworthy observation pertains to the discrepancy in the presence of cloud shadows between the two datasets. In the Biome dataset, cloud shadows are relatively scarce. This can be attributed to two factors: first, not all scenes in the Biome dataset were annotated for cloud shadows, and second, the dataset contains a significant number of scenes that are entirely obscured by clouds, leaving no room for shadow occurrences.
In contrast, the SPARCS dataset lacks a substantial representation of cloudy images. Consequently, evaluating the performance of cloud detection algorithms on scenes entirely shrouded by clouds becomes a challenging endeavor due to the dataset's inherent limitations in this regard.

In the context of evaluating cloud detection algorithms, using the Biome dataset is generally considered more reliable. This preference stems from its larger size and fewer drawbacks compared to the SPARCS dataset. The Biome dataset's subdivision into different biomes further allows for individualized evaluation, enhancing its utility for specific applications. However, it's important to acknowledge that discrepancies in evaluation outcomes are expected when models are trained on one dataset and evaluated on the other. These differences arise due to variations in class and band distributions between the datasets.
Analysts who annotated both datasets exhibit some level of disagreement, introducing an element of subjectivity into the evaluation process. This human factor should be considered when interpreting evaluation results.

Aside from the two sets described previously, other remote satellite images datasets were used only as validation sets and for evaluation purposes.
These are the \href{https://www.kaggle.com/datasets/sorour/95cloud-cloud-segmentation-on-satellite-images}{Cloud 95 dataset} of which only the newest added 75 images were used and the \href{https://mlhub.earth/data/ref_cloud_cover_detection_challenge_v1}{Sentinle-2 dataset}, the latter was discarded due to the poor quality of the cloud masking, also only a subset of 1000 image was used over the full set of around 23000 images.

\section{The Models}\label{state-of-the-art}

\subsection{U-Net Architecture}
\begin{figure}[!h]
    \begin{center}
        \includegraphics[width=\textwidth]{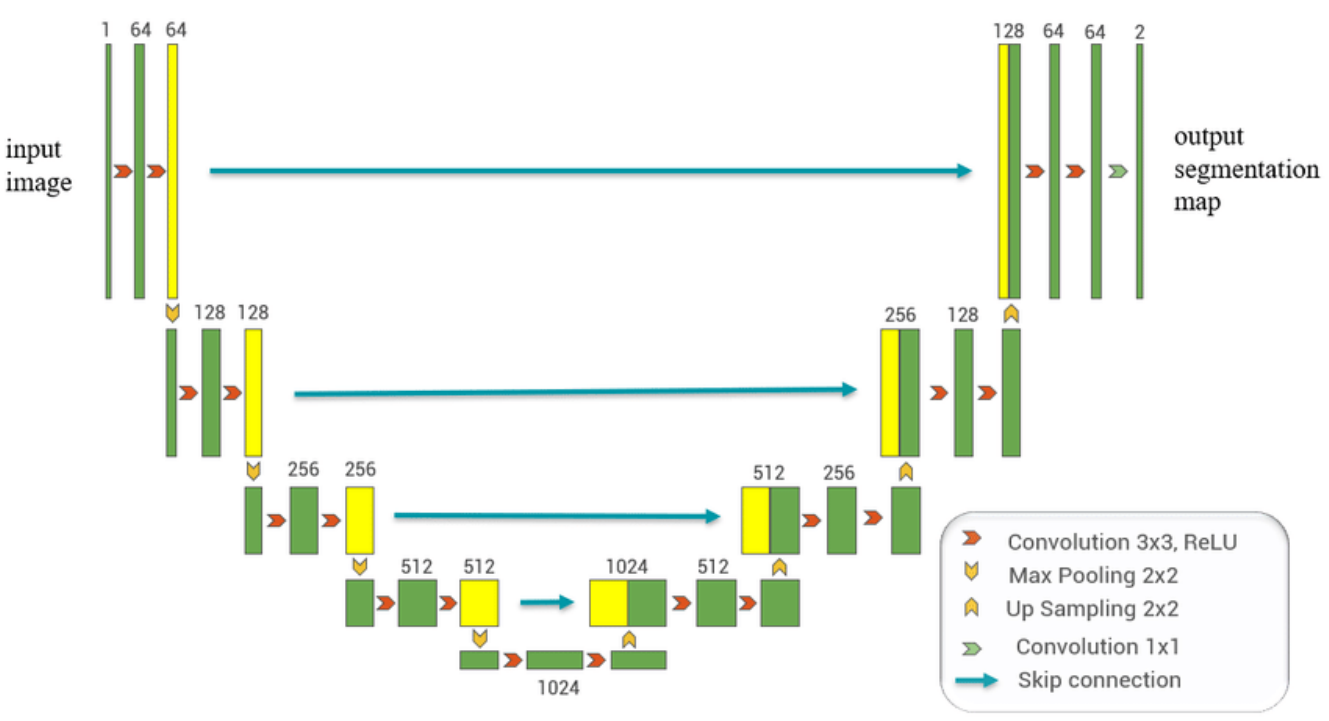}
    \end{center}
    \caption{The encoder-decoder architecture of a standard U-Net \cite{ronneberger_u-net_2015}.}
    \label{fig:unet_standard}
\end{figure}

The U-Net architecture (Fig \ref{fig:unet_standard}), by Ronneberger et al. \cite{ronneberger_u-net_2015} distinguished among convolutional neural networks (CNNs) for its efficiency in generating exact segmentation masks, originally proposed for medical applications such as MRI segmentation \cite{DBLP:journals/corr/abs-1807-10165, ZHOU2021115625, AKTER2024122347}, is characterized by several advantageous features. Notably, its architectural configuration, featuring an encoder-decoder design, excels in the production of dense predictions by sequentially processing image information in a higher-dimensional format. It is noteworthy that U-Net's architectural composition exclusively comprises convolutional layers without the incorporation of dense layers, thereby demonstrating a decent adaptability to images of diverse dimensions and resolutions.

Within the structural framework of U-Net \cite{ronneberger_u-net_2015}, two symmetrical pathways emerge, a contracting path (encoder) and an expanding path (decoder), each consisting of an equivalent number of layers. The encoder, akin to conventional CNNs, serves the purpose of comprehending contextual aspects within an image, extracting meaningful feature maps, and progressively doubling the channel count while halving spatial dimensions. In contrast, the decoder focuses on upsampling feature maps, effecting a twofold increase in spatial dimensions concomitant with a halving of channel numbers. This process, achieved through the application of transposed 2D convolutions in conjunction with regular convolutions, contributes significantly to precise localization and output resolution enhancement.

The construction of the encoder requires a traditional stack of convolution and maxpooling layers \cite{boureau2010theoretical} that systematically increase the receptive field and iteratively reduce the spatial resolution through the use of two 3 × 3 filters and a 2 × 2 maxpooling layer. This process is repeated four times, with each iteration doubling the number of feature channels. Although this approach is good at capturing “WHAT” information in an image, it tends to interfere with retaining the “WHERE” information. The complementary symmetric decoder \cite{sym13122246}, a central component of the architecture, uses transposed convolutional layers for data upsampling, with a strategic focus on maintaining final dimensions that are very similar to those of the input images.

One significant characteristic of U-Net architectures is the integration of "symmetric skip connections" \cite{mao2016image}. These connections establish a U-shaped architecture and allow feature maps to be concatenated from corresponding layers in the encoder and decoder. This symmetrical connection mechanism significantly improves pixel-level localization accuracy and provides accurate location details that prove critical in supporting segmentation efforts. Following each concatenation, standard convolutions are applied, providing the model with opportunities for further refinement in output generation.

The distinctive U-Net architecture, marked by encoder-decoder symmetry and the strategic incorporation of symmetric skip connections, underscores its exceptional performance in precise segmentation tasks, particularly in scenarios characterized by limited training data. The U-shaped design, combined with its convolutional layer exclusivity and skip connection strategy, imparts a notable degree of adaptability to diverse image dimensions. Additionally, the model's proficiency in maintaining segmentation accuracy across varying input sizes further underscores its utility in practical applications beyond its original medical context.

\subsection{U-Net++ - an improved U-Net architecture}
\begin{figure}[H]
    \begin{center}
        \includegraphics[width=0.7\textwidth]{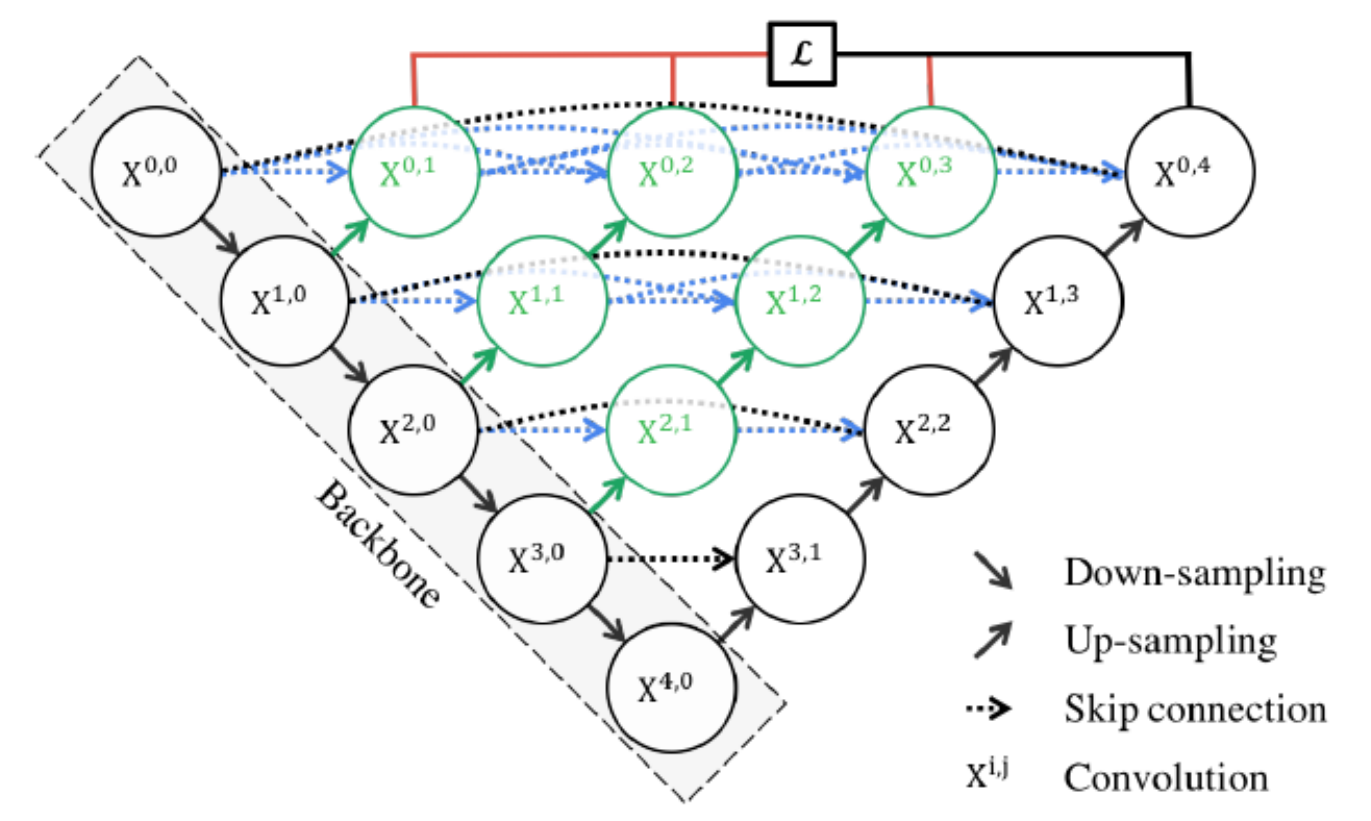}
    \end{center}
    \caption{The U-Net++ architecture from \cite{zhou_unet_2018}.}
    \label{fig:unet_plusplus}
\end{figure}

U-Net++, introduced in the seminal work by Zhou et al. \cite{zhou_unet_2018}, represents a refined iteration of the original U-Net architecture, incorporating advancements primarily centred around redesigned nested and dense skip pathways. These modifications, visually depicted in green and blue within Figure \ref{fig:unet_plusplus}, are complemented by the integration of deep supervision elements, represented in red. The fundamental premise underlying these enhancements is the efficient bridging of the semantic gap existing between the encoder and decoder feature maps. This bridging process unfolds through the progressive enrichment of fine-grained feature maps \cite{hariharan2016object} derived from the encoder subnetwork, culminating in their merger with corresponding feature maps from the decoder subnetwork.

As illustrated in Figure \ref{fig:unet_plusplus}, the architecture of U-Net++ is characterized by its incorporation of dense skip connections, a concept inspired by DenseNet \cite{Huang_2017_CVPR}. The objective here is to augment gradient flow and foster enhanced information transfer between network layers. This augmentation transpires within dense convolution blocks, each consisting of multiple convolution layers. For instance, the skip pathway connecting nodes \(X^{1,0}\) and \(X^{1,3}\) exemplifies a dense convolution block housing two convolution layers. Within such a dense block, each convolution layer receives input from its predecessor within the same block, and its output is concatenated with the results of preceding convolution layers. The effectiveness of this architectural augmentation in addressing the task at hand is corroborated by empirical evidence presented in \cite{kaur_buttar_semantic_2022}.

Expanding upon the U-Net++ model detailed in the \cite{zhou_unet_2018}, the authors propose a nested U-Net framework designed to optimize medical image segmentation. The nested architecture introduces redesigned skip pathways, where both nested (green) and dense (blue) connections play pivotal roles. The deep supervision mechanism (red) facilitates effective information transfer between the encoder and decoder feature maps, mitigating the semantic gap challenge.

\subsection{RS-net Architecture}

RS-Net, as presented in the study by Jeppesen et al. \cite{jeppesen_cloud_2019}, constitutes an advanced deep learning model rooted in the U-net architecture \cite{ronneberger_u-net_2015}, specifically crafted for the task of cloud classification in the domain of remote sensing. The significance of spatial patterns in cloud detection algorithms is underscored in the investigation, with RS-Net effectively incorporating these patterns to achieve heightened performance compared to conventional methodologies \cite{jeppesen_cloud_2019}. 

The evaluation encompasses the Landsat 8 dataset \cite{foga_cloud_2017}, in which RS-Net is systematically compared against the Fmask algorithm \cite{zhu_object-based_2012}. Notably, the paper delves into the superior performance demonstrated by an RS-Net model trained on annotations generated by the Fmask algorithm. The processing time for classification is reported as 18.0 ± 2.4 seconds per Landsat 8 product on a standard consumer PC.

The core architectural tenets of RS-Net, rooted in a fully convolutional network tailored for pixel-wise classification tasks in remote sensing applications, aim to map an input image (X) to a pixel-wise classification map (Y), providing a confidence metric for cloud presence. The model's composition involves multiple layers, encompassing convolutional layers, max-pooling layers, and upsampling layers, ultimately encapsulating a composition of two functions (f1 and f2).

During the training regimen, the utilization of binary cross-entropy loss facilitates the comparison between the predicted cloud mask (Y) and the true cloud mask ($\hat{Y}$). The back-propagation algorithm is employed to compute gradients for each parameter, facilitating optimization through the application of gradient descent. The architecture integrates various regularization techniques, including L2-regularization, dropout, and batch normalization layers, to mitigate the risk of overfitting. Furthermore, a cropping layer is seamlessly incorporated to eliminate underperforming regions proximal to the image borders.

\begin{figure}[h]
    \begin{center}
        \includegraphics[width=1.0\textwidth]{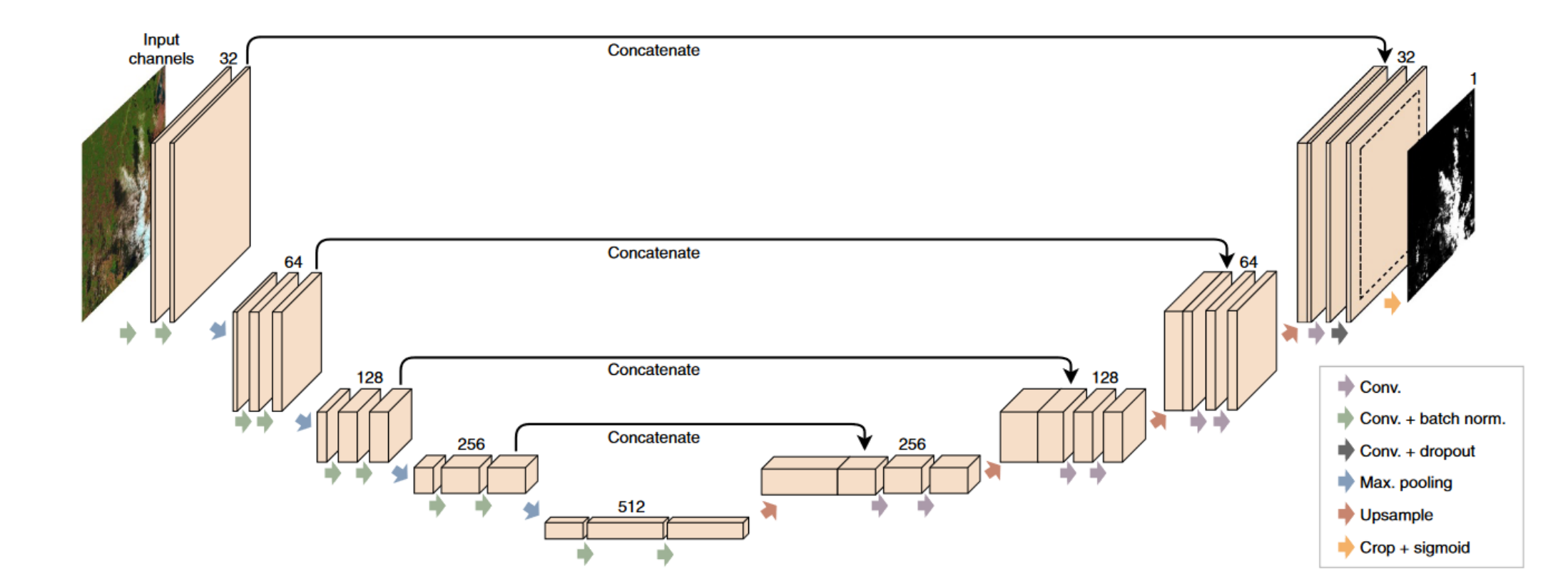}
    \end{center}
    \caption{The RS-Net architecture, based on U-net(\ref{fig:unet_standard}), where the depth of the feature maps in the stages are provided by the number above each stage.}
    \label{fig:rsnet_a}
\end{figure}

The RS-Net architecture, as depicted in Figure \ref{fig:rsnet_a}, adheres to the U-net design paradigm, comprising an encoder and decoder. The encoder, fashioned with convolutional and max-pooling layers \cite{DBLP:journals/corr/abs-2009-07485}, serves the purpose of downsampling feature maps, while the decoder, equipped with skip connections \cite{DBLP:journals/corr/DrozdzalVCKP16}, undertakes the task of upsampling maps to offset information loss incurred during max-pooling.

The model exhibits adaptability, permitting training on diverse spectral bands, and enabling a comprehensive exploration of its performance across various band combinations. The evaluation in the paper encompasses five distinct band combinations, leveraging Keras with TensorFlow as the backend.

In the course of training, input data normalization ensues, scaling values between 0 and 1. Image patches are employed for classification, and a clipping layer is introduced to address border-related challenges. Data augmentation strategies, including horizontal and vertical flips, contribute to the expansion of the training dataset. The AMSGrad\cite{reddi_convergence_2019} variant of the Adam optimizer\cite{kingma_adam_2017} is deployed, and hyperparameter optimization is systematically undertaken through a combination of grid search and random search methodologies.

\subsection{YOLOv8 segmentation}

You Only Look Once version 8 (YOLOv8), an algorithm for real-time object detection and classification, represents a significant advancement developed by Joseph Redmon and Ali Farhadi. Building upon its predecessors, YOLOv8 \cite{reis_real-time_2023} introduces an improved precision and swiftness, positioning itself as a preferred choice for applications spanning autonomous vehicles, security systems, and video analysis specially for object detection operations \cite{WANG2022116793, ZHOU2024122256}.

\begin{figure}[!h]
    \begin{center}
        \includegraphics[width=\textwidth]{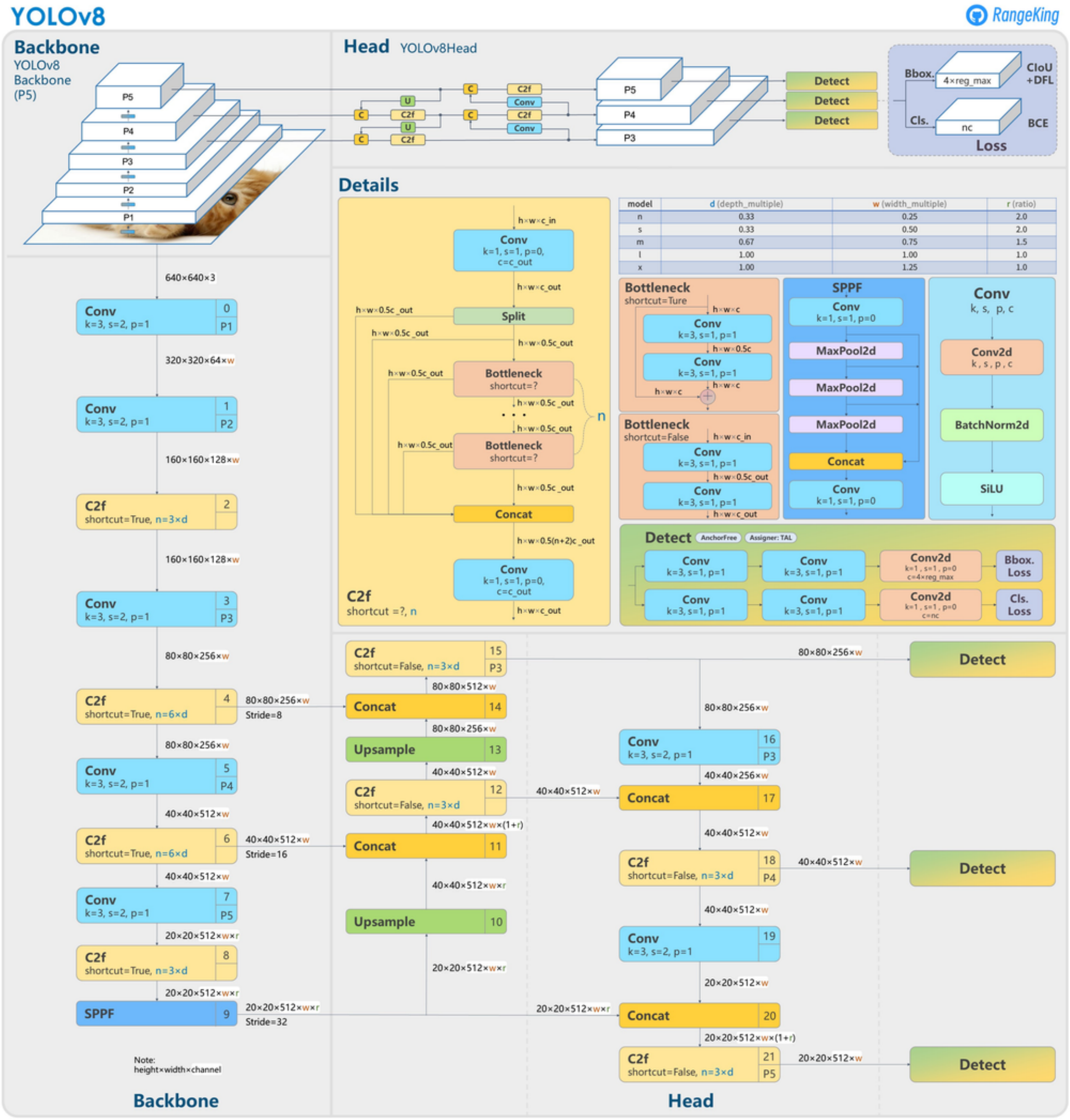}
    \end{center}
    \caption{Detailed architecture of latest You Only Look Once version (YOLOv8) \cite{reis_real-time_2023}.}
    \label{fig:yolov8}
\end{figure}

YOLOv8 comprises three key components \cite{reis_real-time_2023, photon_revolutionizing_2023}:
\begin{itemize}
    \item \textbf{Backbone Network}: Serving as the foundation, this component utilizes a ResNet-50 deep convolutional neural network (CNN) to extract features from input images. Trained on extensive datasets like ImageNet, the CNN excels in recognizing high-level features in images.
    
    \item \textbf{Neck Network}: The intermediary between the backbone network and the head network, this component plays a pivotal role in reducing feature map size while enhancing feature resolution. Striking a balance between accuracy and speed, the neck network contributes to YOLOv8's efficiency.
    
    \item \textbf{Head Network}: Responsible for predicting object class and location, this component utilizes anchor boxes to estimate object locations and class scores. The head network undergoes training to optimize Intersection Over Union (IoU) between predicted and ground-truth boxes. Additionally, it integrates the non-maximum suppression (NMS) algorithm, filtering out overlapping bounding boxes and ensuring the output comprises only the most confident predictions.
\end{itemize}

Versatility characterizes YOLOv8, as it accommodates a spectrum of tasks, including detection, instance segmentation, pose/keypoints, and classification. For our specific application, we employed the instance segmentation configuration. To tailor it to our task, a post-processing step was undertaken on the predictions, transitioning from instance segmentation (outputting a mask for each detected cloud instance) to semantic segmentation (outputting a single mask encompassing all detected clouds).

\subsection{Google DeepLab V3+}

DeepLab is a series of models designed for image semantic segmentation, with its latest iteration by Chen et al.\cite{chen2018encoderdecoder, chen2017rethinking, chen2017deeplab, chen2016semantic}, namely v3+, establishing itself as state-of-the-art. Notable among its advancements is the incorporation of atrous spatial pyramid pooling (ASPP), a semantic segmentation architecture that builds upon DeepLabv2 by Chen et al.\cite{chen2017deeplab} with several enhancements. These modifications aim to address the challenge of segmenting objects at multiple scales by introducing modules that employ atrous convolution either in cascade or in parallel. Furthermore, the ASPP module from DeepLabv2 is enhanced by integrating image-level features, thereby capturing global context and enhancing overall performance.

A pivotal modification to the ASPP module involves the application of global average pooling on the final feature map of the model. The resulting image-level features undergo processing through a 1×1 convolution with 256 filters (and batch normalization) before bilinear upsampling to the desired spatial dimensions. The refined ASPP module now comprises one 1×1 convolution and three 3×3 convolutions with rates = (6, 12, 18) when the output stride is set to 16, each with 256 filters and batch normalization, in addition to the image-level features.

In its v3+ iteration, the model introduces a straightforward yet efficient decoder module. This extension incorporates an encoder-decoder architecture, proven to be beneficial in various literature, including FPN\cite{lin2017feature}, DSSD\cite{fu2017dssd}, TDM\cite{shrivastava2017skip}, SharpMask\cite{pinheiro2016learning}, RED-Net\cite{jiang2018rednet}, and U-Net\cite{zhou_unet_2018}, for diverse purposes.

\begin{figure}[H]
    \begin{center}
        \includegraphics[width=0.8\textwidth]{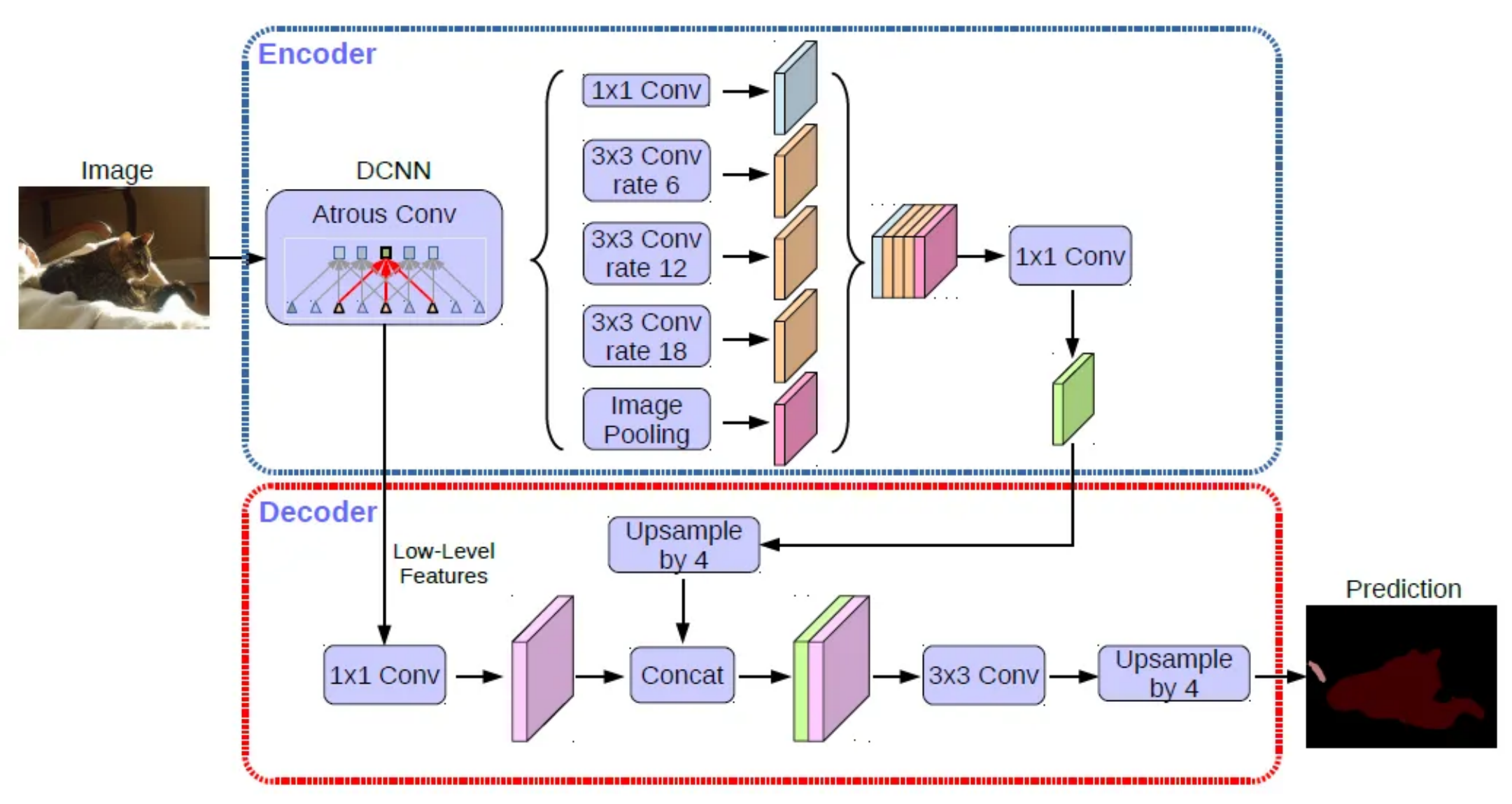}
    \end{center}
    \caption{DeepLabv3+ architecture reproduced from \cite{chen2018encoderdecoder}.}
    \label{fig:dlv3a}
\end{figure}

ASPP plays a crucial role in obtaining multi-scale context information, and prediction results are obtained through upsampling. The ASPP network, applied on the feature map extracted from the backbone, employs four parallel atrous convolutions with different atrous rates to handle object segmentation at various scales. Image-level features are also incorporated to capture global context information. Post all operations, the results from each operation are concatenated along the channel, and a 1×1 convolution is applied to generate the final output.

To enhance the network's speed and robustness, Modified Aligned Xception, and Atrous Separable Convolution are introduced\cite{chen2018encoderdecoder}. Atrous convolution \cite{mpdis21217191}, also known as dilated convolution, is a pivotal component within the DeepLabv3+ architecture. This convolutional operation introduces a controllable receptive field, allowing for the integration of contextual information at different scales without the need for downsampling. The atrous convolution is applied over the input feature map, where the atrous rate corresponds to the stride with which the input signal is sampled. This facilitates the extraction of features at varying scales, contributing to the network's ability to handle objects of diverse sizes.

The Atrous Separable Convolution combines the efficiency of depthwise separable convolution with the adaptability of atrous convolution. This factorization of a standard convolution into a depthwise convolution followed by a point-wise convolution (1×1 convolution) significantly reduces computational complexity. This approach offers a computational advantage while maintaining or even improving the overall performance of the model. They are particularly effective in scenarios where computational efficiency is crucial, as seen in models like MobileNetV1\cite{howard2017mobilenets}.

The Encoder-Decoder Architecture forms a fundamental framework within DeepLabv3+, playing a pivotal role in tasks such as image classification and semantic segmentation. For image classification, the spatial resolution of the final feature maps is typically 32 times smaller than the input image resolution, resulting in an output stride of 32. However, for semantic segmentation tasks, this output stride is often deemed too small. To address this, DeepLabv3+ allows for the adjustment of the output stride to 16 (or 8), facilitating denser feature extraction by removing striding in the last one (or two) blocks and applying atrous convolution correspondingly.

In the encoder stage, convolutional features are extracted and processed through the ASPP module, which probes convolutional features at multiple scales by applying atrous convolution with different rates. Image-level features are also incorporated to provide global context information. The decoder stage involves bilinear upsampling of the encoder features by a factor of 4, followed by concatenation with corresponding low-level features. To manage the channel complexities, a 1×1 convolution is applied on low-level features before concatenation. Subsequently, a series of 3×3 convolutions refines the features, followed by a simple bilinear upsampling by a factor of 4.

Figure \ref{fig:dlv3a} provides a visual representation of the DeepLabv3+\cite{chen2018encoderdecoder} architecture, emphasizing the interaction between the encoder and decoder components. This intricate interplay of Atrous Convolution, Atrous Separable Convolution, and the Encoder-Decoder Architecture contributes to the model's ability to capture intricate details and contextual information, making it highly effective for semantic segmentation tasks.

\subsection{Cloud-Net}

The model architecture proposed in \cite{mohajerani_cloud-net_2019} employs a convolutional neural network (CNN) with dual branches \cite{lu2022dual}, namely the contracting arm and the expanding arm, to obtain a cloud mask at the output with the same dimensions as the input image. The contracting arm is tasked with feature extraction, generating deep low-level features from the input image, while the expanding arm uses these features to discern cloud attributes, recover them, and ultimately generate the output cloud mask. This output is presented as a probability map, where each pixel's value signifies the likelihood of belonging to the cloud class.

\begin{figure}[h]
    \begin{center}
        \includegraphics[width=0.7\textwidth]{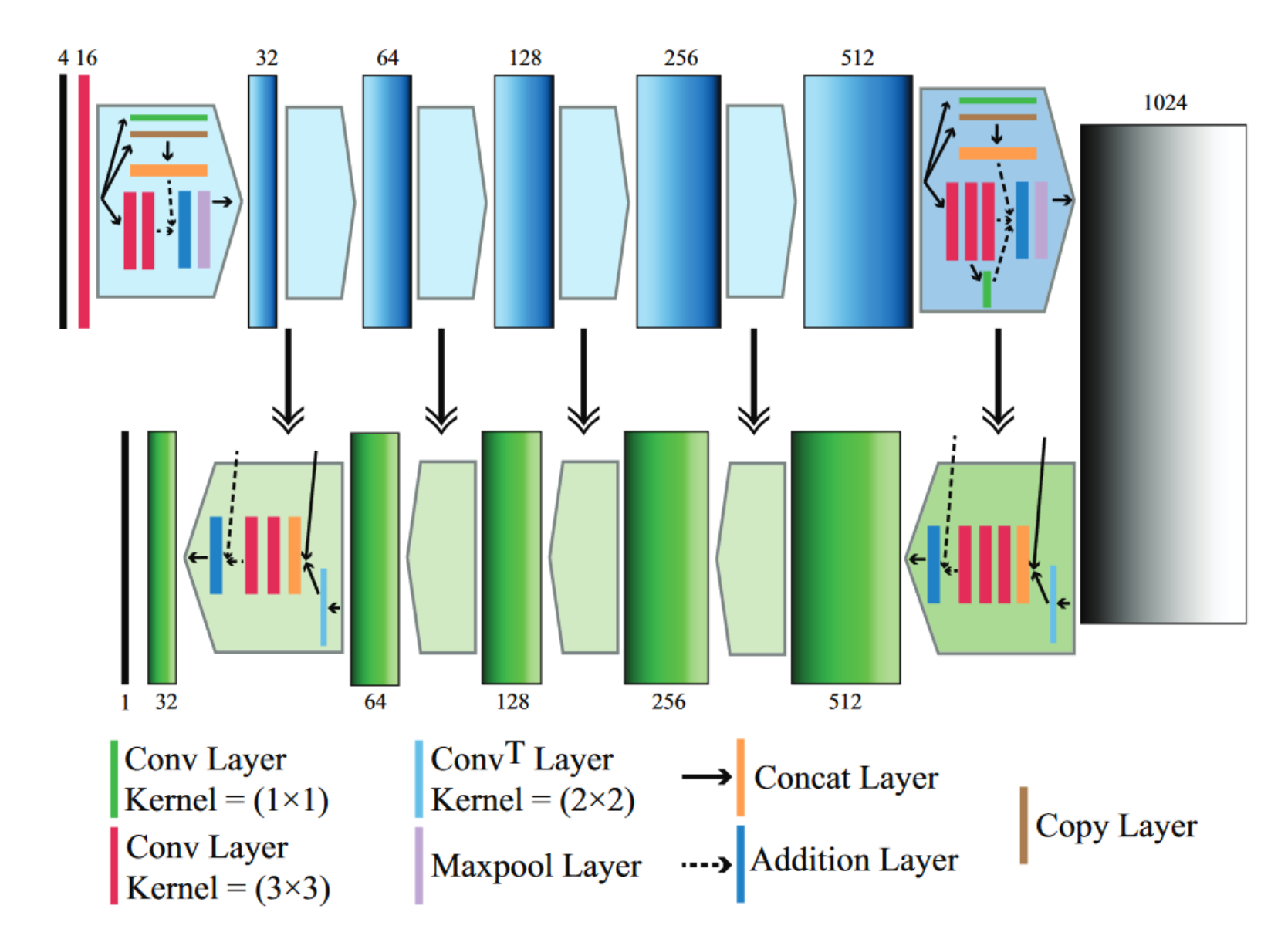}
    \end{center}
    \caption{Cloud-Net architecture. ConvT, Concat, and Maxpool refer to convolution transposed, concatenation, and maxpooling, respectively. The bars with gradient shading represent the feature maps. The numbers on the top and the bottom of the bars are the corresponding depth of each feature map. Reproduced from \cite{mohajerani_cloud-net_2019}}
    \label{fig:cloudnet}
\end{figure}

The Cloud-Net, a cloud segmentation CNN, follows this dual-branch architecture, as illustrated in Figure \ref{fig:cloudnet}. The contracting arm is represented by the blue bars and blocks, responsible for feature extraction, while the green arrows and blocks constitute the expanding arm, tasked with attribute retrieval and cloud mask generation. The inclusion of shortcut connections in each block, encompassing Concat, Copy, and Addition layers (depicted in Figure \ref{fig:cloudnet}), facilitates the preservation and utilization of learned contexts from earlier layers. This design choice enhances the network's capability to capture diverse cloud features. Furthermore, these shortcut connections contribute to expediting the training process by mitigating the vanishing gradient phenomenon during backpropagation.

The interconnections between the two arms of Cloud-Net play a pivotal role in enhancing the accuracy of the generated cloud mask by the expanding arm. Common geometric data augmentation techniques, such as horizontal flipping, rotation, and zooming, are applied before each training epoch to augment the training dataset. The activation function used in the convolution layers of Cloud-Net is Rectified Linear Unit (ReLU) \cite{nair_rectified_nodate}, while a sigmoid layer is employed in the final convolution layer of the network.

\subsection{CloudX-Net}

The architectural framework introduced by Kanu et al. \cite{kanu_cloudx-net_2020}, known as CloudX-Net, builds upon the foundation laid by the CloudNet model\cite{mohajerani_cloud-net_2019}. CloudNet, designed to capture both global and local features of input for cloud segmentation, demonstrates proficiency in generating cloud masks. However, when benchmarked against dedicated cloud segmentation architectures like RS-Net, it falls short of optimal performance.

To overcome this limitation, CloudX-Net integrates Atrous Spatial Pyramid Pooling (ASPP)\cite{chen2018encoderdecoder} and Separable Convolution, elucidated in the Google DeepLab segment, as influential enhancements. ASPP plays a pivotal role in capturing multi-spectral spatial features, thereby improving classification accuracy. Simultaneously, Separable Convolution reduces the number of parameters needed for training, enhancing the efficiency of the deep learning-based cloud detection method.

Notably, CloudX-Net exhibits superior performance across various evaluation metrics compared to existing deep learning architectures, with the exception of Recall. The model excels in finely segmenting clouds, surpassing other models that predict only the coarse outline of clouds convincingly. Moreover, the reduced number of parameters not only mitigates the risk of overfitting but also minimizes storage requirements.

Quantitative evaluation on both Landsat 8 image dataset and the 38-Cloud Dataset underscores the effectiveness of CloudX-Net. The model demonstrates a noteworthy 0.76\% improvement in the Adjusted Jaccard Index (AJI) on the Landsat 8 dataset and a more substantial improvement of 2.06\% on the 38-Cloud Dataset. These results affirm the model's efficacy in significantly enhancing the accuracy of cloud detection through segmentation.

In conclusion, the CloudX-Net architecture represents a substantial advancement over the CloudNet model, incorporating influential tools that not only augment the efficiency and precision of the architecture in cloud detection through segmentation but also yield tangible improvements in performance metrics.

\section{Benchmark}\label{bechmark}

\subsection{Setup}

The table \ref{software-setup} provides an overview of the essential software packages used in the paper. The input image size was fixed at 256 × 256 pixels(since some architectures need to work with fixed sizes). The GPU memory in some cases has been a limiting factor with batch sizes larger than 8.
For the training and the evaluation of the model 2 NVIDIA GeForce RTX 3090 with 24G of memory have been utilized. These processors were made available by the \href{https://www.supsi.ch/en/home}{University of Applied Sciences and Arts of Southern Switzerland} and the \href{https://www.idsia.ch/}{IDSIA institute}.

\begin{table}[H]
\centering
\caption{\label{software-setup}Software setup}
    \begin{adjustbox}{width=440 pt}
        \begin{threeparttable}[b]
            \begin{tabular}{lll}
                \hline
                Name & Version & Description \\ \hline
                Ubuntu & 20.04.1 LTS & Linux operating system \\
                Python & 3.9 & Programming language \\
                Keras & 2.9 & High-level API used for TensorFlow \\
                TensorFlow & 2.9 & Deep learning framework by Google \\
                CUDA & 11.2 & Platform for GPU based processing (used by TensorFlow) \\
                CUDnn & 8.1 & CUDA library for deep neural networks (used by TensorFlow/CUDA) \\
            \end{tabular}
        \end{threeparttable}
    \end{adjustbox}
\end{table}

For the training process, all models under consideration underwent training using both the SPARCS and Biome datasets, specifically utilizing patches extracted from the images of these datasets. These patches possess dimensions of 256x256 pixels, with a 64-pixel overlap with neighbouring images. This overlap was deliberately selected to ensure that critical spatial information was not lost during the patch creation process. All the images were standardized to values between 0 and 1. The division of the dataset into training, testing, and validation subsets followed an \textbf{80\%, 10\%, 10\%} ratio. Importantly, this division was carried out prior to generating image patches. This precautionary measure was taken to prevent any patches from the same image appearing in both the training and test sets, which could lead to data leakage.

The unique categorization within the Biome dataset facilitated a specific data split. Images were initially grouped by background biome before undergoing the split process. Consequently, each data split maintained a consistent proportion of images with the same background biome type.
By creating these patches, we significantly expanded the dataset size. Specifically, for the Biome dataset, we obtained 17,389 training images, 1,933 validation images, and 3,714 test images. For the SPARCS dataset, the result was 992 training images, 128 validation images, and 128 test images.
While exploring the training process, various data augmentation techniques were tested. However, they did not yield significant improvements, so they were not integrated. As an example, one such augmentation technique involved introducing rotations, variations in image brightness or flips with certain probabilities for each image.

\subsection{Results}
\subsubsection{Biome-based training}: To compare the deep learning models, AUC (Area Under the Curve), Dice, IoU (Intersection over Union), and coverage similarity metrics have been selected. Those metrics are commonly used in the evaluation of segmentation models \cite{Mller2022TowardsAG, Basu2023}. AUC \cite{BRADLEY19971145} measures the performance of a binary classification model by calculating the area under the ROC curve. Dice \cite{ZOU2004178} and IoU \cite{Rahmaniou2016} are similarity measures that compare the overlap between the predicted segmentation and the ground truth segmentation. Coverage similarity measures the percentage of the ground truth segmentation that is covered by the predicted segmentation. These metrics provide a quantitative way to evaluate the accuracy and effectiveness of segmentation models and are widely used in various fields such as medical imaging, computer vision and remote sensing.

Table \ref{results_biome} provides the comparison of the performances of the selected 
7 segmentation models, namely U-Net, RS-Net, U-Net++, DeepLabV3+, Cloud-Net, and YOLOv8 when trained on the Biome dataset and evaluated on three different datasets: SPARCS, Biome, and 95 Clouds. 

\begin{itemize}
    \item Across all three datasets, DeepLabV3+ emerges as a strong-performing model, consistently ranking high in all four evaluation metrics. It maintains a mean of 0.92 and 0.82 respectively for the instances trained on Biome and SPARCS, the IoU values being higher than 0.80 in 5 out of 6 tests confirm the accuracy and precision of the segmentation. These results indicate that DeepLabV3+ exhibits robust performance in segmenting various types of clouds, including cumulus, stratus, and cirrus clouds, as well as distinguishing between cloudy and clear regions.

    \item RS-Net also demonstrates strong performance across the three datasets, particularly in terms of AUC and coverage similarity. It achieves an AUC score of 0.9036 on SPARCS, 0.9232 on Biome, and 0.7933 on 95 Clouds. In terms of coverage similarity scores an accuracy of 0.9879 has been detected when tested on SPARCS, 0.9705 on Biome, and 0.9797 on 95 Clouds. However, its Dice and IoU scores are generally lower than those of DeepLabV3+, suggesting that RS-Net may struggle slightly more in accurately delineating the boundaries of cloudy regions.
    
    \item Cloud-Net shows competitive performance on all three datasets, especially in terms of Dice and IoU scores. It achieves Dice scores of 0.8676 on SPARCS, 0.9058 on Biome, and 0.7499 on 95 Clouds, and IoU scores of 0.7662 on SPARCS, 0.8278 on Biome, and 0.5999 on 95 Clouds. Its AUC and coverage similarity scores are somewhat lower than those of DeepLabV3+ and RS-Net, but still indicative of good performance.
    
    \item U-Net and U-Net++ tend to perform relatively poorly compared to the other models, particularly in terms of AUC and coverage similarity. U-Net achieves AUC scores of 0.8914 on SPARCS, 0.8844 on Biome, and 0.7573 on 95 Clouds, while U-Net++ achieves AUC scores of 0.7336 on SPARCS, 0.7399 on Biome, and 0.7132 on 95 Clouds. Their Dice and IoU scores are also generally lower than those of the top-performing models.
    
    \item Lastly, YOLOv8 tends to perform poorly across all three datasets, with low scores in all four evaluation metrics. Its IoU scores are similarly low, at 0.2079 on SPARCS, 0.1331 on Biome, and 0.0863 on 95 Clouds.
\end{itemize}

\begin{table}[H]
\centering
\caption{Evaluation results for the Biome trained models.}
\label{results_biome}
    \begin{adjustbox}{width=260 pt}
        \begin{threeparttable}[b]
            \small 
            \begin{tabular}{lcccccccc}
                \hline
                Model & AUC & Dice & IoU & Coverage similarity \\ \Xhline{2\arrayrulewidth}
                \multicolumn{9}{l}{\textbf{Trained on Biome ground truth and tested on SPARCS}} \\
                U-Net & 0.8914 & 0.7964 & 0.6617 & 0.9879 \\
                RS-Net & 0.9036 & 0.8415 & 0.7264 & \textbf{0.9972} \\
                U-Net++ & 0.7336 & 0.4167 & 0.2632 & 0.6420 \\
                DeepLabV3+ & 0.9052 & \textbf{0.8688} & \textbf{0.7680} & 0.9841 \\
                Cloud-Net & \textbf{0.9053} & 0.8676 & 0.7662 & 0.9847 \\
                CloudXNet & 0.8112 & 0.5588 & 0.3878 & 0.8548 \\
                YOLOv8 & 0.6042 & 0.3442 & 0.2079 & 0.8828 \\ \Xhline{2\arrayrulewidth}
                \multicolumn{9}{l}{\textbf{Trained on Biome ground truth and tested on Biome}} \\
                U-Net & 0.8844 & 0.9038 & 0.8245 & 0.9498 \\
                RS-Net & 0.9232 & 0.9346 & 0.8772 & 0.9705 \\
                U-Net++ & 0.7399 & 0.8157 & 0.6887 & 0.7962 \\
                DeepLabV3+ & \textbf{0.9348} & \textbf{0.9361} & \textbf{0.8798} & 0.9668 \\
                Cloud-Net & 0.8994 & 0.9058 & 0.8278 & \textbf{0.9831} \\
                CloudXNet & 0.8936 & 0.8906 & 0.8028 & 0.9304 \\
                YOLOv8 & 0.5643 & 0.235 & 0.1331 & 0.5291 \\ \Xhline{2\arrayrulewidth}
                \multicolumn{9}{l}{\textbf{Trained on Biome ground truth and tested on 95 Clouds}} \\
                U-Net & 0.7573 & 0.7633 & 0.6172 & 0.9104 \\
                RS-Net & 0.7933 & \textbf{0.7796} & \textbf{0.6388} & \textbf{0.9797} \\
                U-Net++ & 0.7132 & 0.7285 & 0.5729 & 0.8751 \\
                DeepLabV3+ & \textbf{0.8128} & 0.7751 & 0.6328 & 0.8455 \\
                Cloud-Net & 0.7920 & 0.7499 & 0.5999 & 0.8492 \\
                CloudXNet & 0.6861 & 0.6205 & 0.4498 & 0.8553 \\
                YOLOv8 & 0.5394 & 0.1588 & 0.0863 & 0.5726 \\
                \hline
            \end{tabular}
        \end{threeparttable}
    \end{adjustbox}
\end{table}

The results presented in Fig.\ref{fig:b_loss} demonstrate the loss functions of the seven segmentation models. While YOLO exhibits relatively stable loss curves, this observation may be somewhat misleading as the asymptote is attained at approximately 0.9 during training and 1.2 during testing. Conversely, the remaining models appear to exhibit greater variability; however, this is largely attributable to their lower loss values, which stabilize at an asymptote of roughly 0.05 to 0.35 by epoch 100. Furthermore, RS-Net demonstrates superior stability among these models. Note that this series of models was trained with at most 100 epochs, given the very slow training time and also because when using an early-stopping function the algorithms stopped learning at around 100 epochs. The only exception is YOLO which uses built-in functions.

\begin{figure}[h!]
    \centering
\begin{subfigure}{6cm}
    \includegraphics[width=\textwidth]{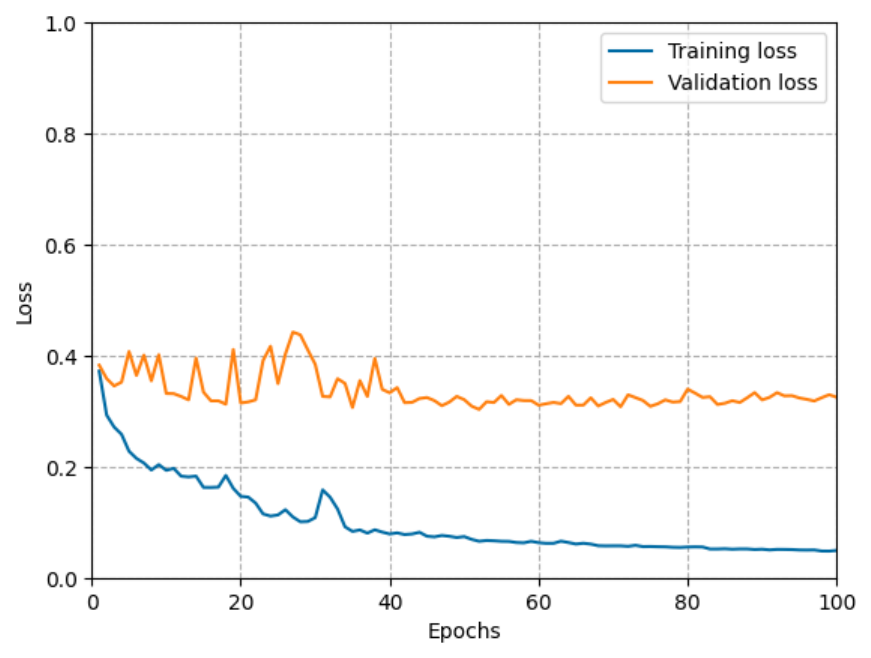}
    \caption{U-Net.}
    \label{fig:firstb}
\end{subfigure}
\hfill
\begin{subfigure}{6cm}
    \centering
    \includegraphics[width=\textwidth]{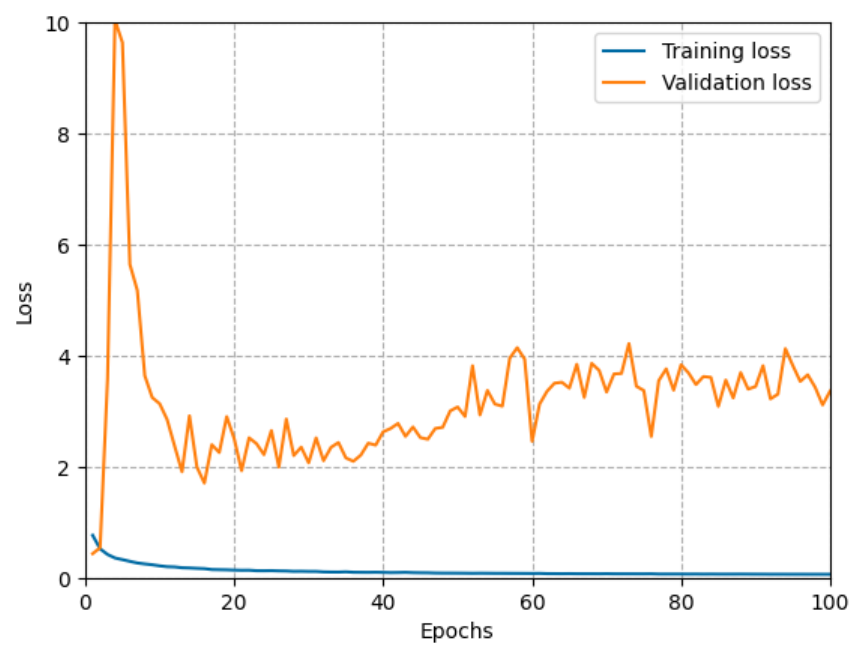}
    \caption{U-net++.}
    \label{fig:secondb}
\end{subfigure}
\caption{The two U-Net iterations in \ref{fig:firstb} and \ref{fig:secondb} show bad stability, however, while the U-Net++ version overfit on the training, the standard version of the model has a lower gap between training and testing curves.}
\label{fig:b_loss}
\end{figure}
\begin{figure}[h!]
\ContinuedFloat
\begin{subfigure}{6cm}
    \centering
    \includegraphics[width=\textwidth]{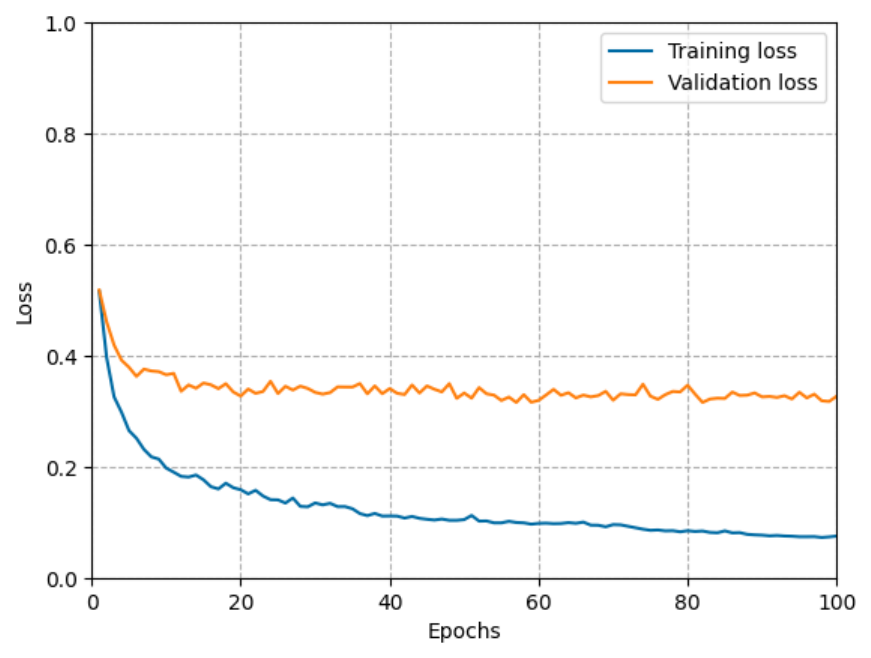}
    \caption{RS-Net.}
    \label{fig:thirdb}
\end{subfigure}
\hfill
\begin{subfigure}{6cm}
    \centering
    \includegraphics[width=\textwidth]{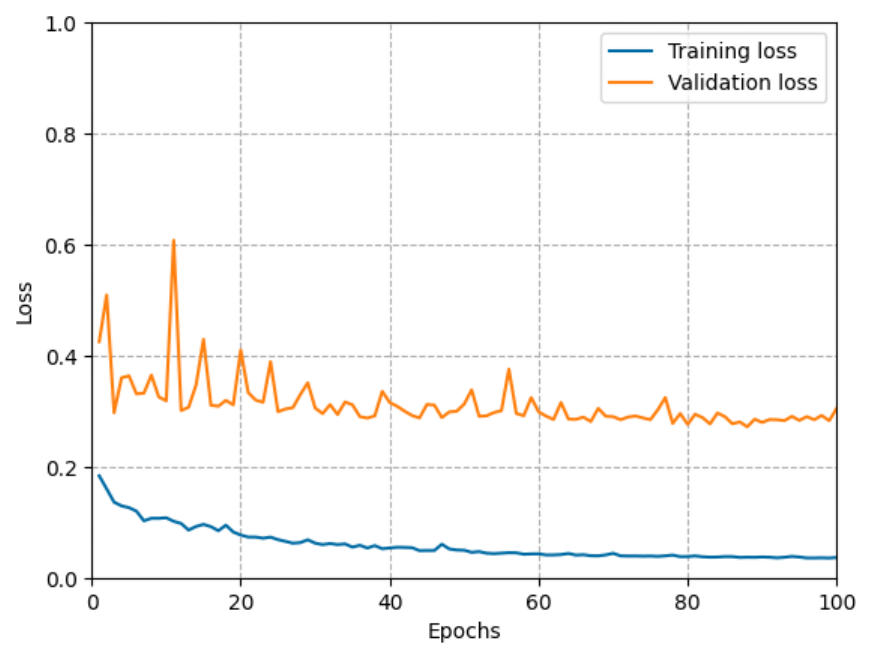}
    \caption{DeepLab V3+.}
    \label{fig:fourthb}
\end{subfigure}    
\caption{Resulting curves of RS-Net \ref{fig:thirdb} and DeepLabV3+ \ref{fig:fourthb} training. The results shows better optimization than the U-Net and U-Net++ as they are more stable and achieve slightly better final values.}
\label{fig:b_loss_}
\end{figure}
\begin{figure}[H]
\ContinuedFloat
\centering
\begin{subfigure}{6cm}
    \centering
    \includegraphics[width=\textwidth]{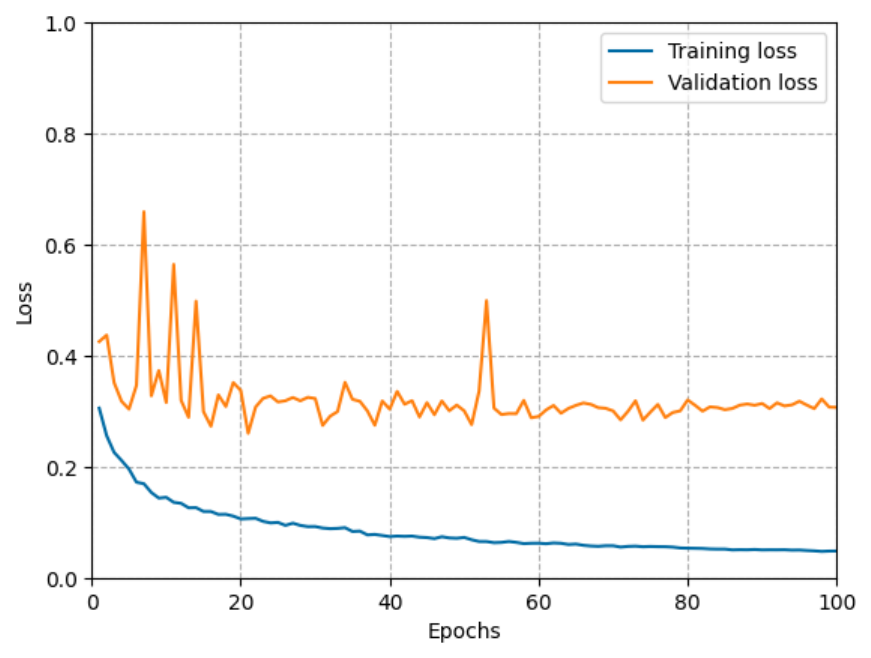}
    \caption{Cloud Net.}
    \label{fig:fifthb}
\end{subfigure}
\hfill
\begin{subfigure}{6cm}
    \centering
    \includegraphics[width=\textwidth]{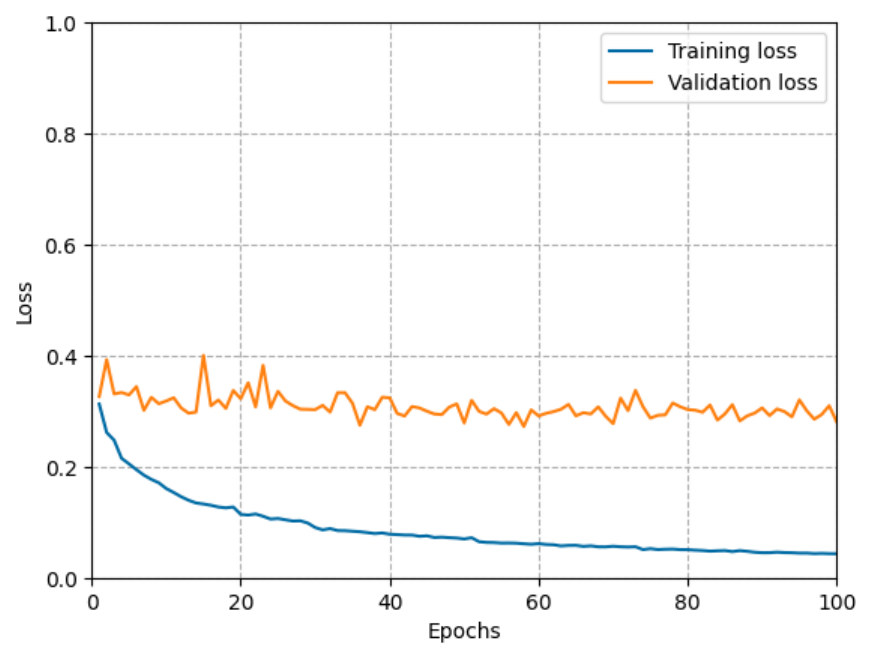}
    \caption{Cloud X Net.}
    \label{fig:sixthb}
\end{subfigure}
\hfill
\begin{subfigure}{6cm}
    \centering
    \includegraphics[width=\textwidth]{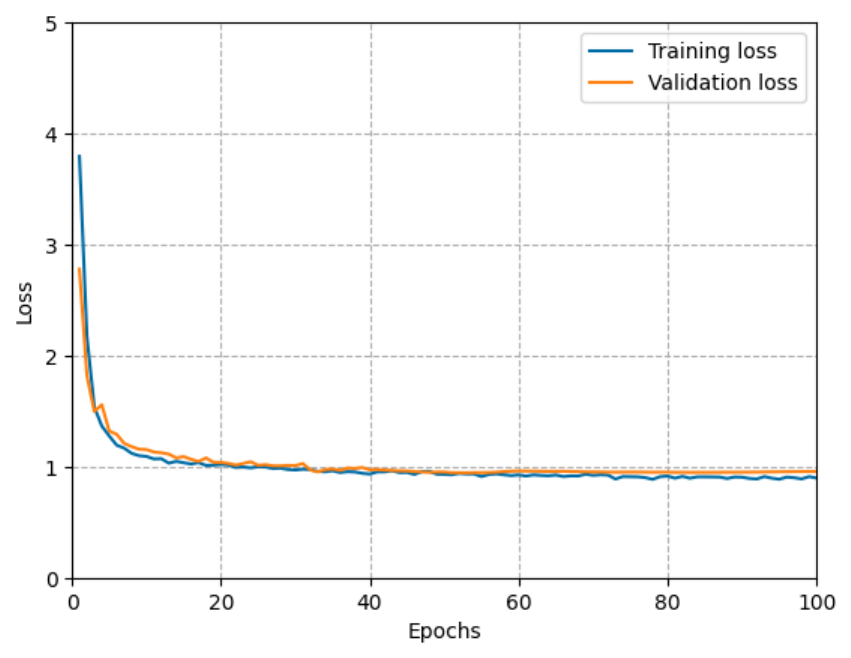}
    \caption{YOLO V8.}
    \label{fig:seventhb}
\end{subfigure}
\caption{The last models have mixed results, as seen in the evaluations YOLO \ref{fig:seventhb} seems stable but does not reach good performances. The CloudNet model \ref{fig:fifthb} has similar stability with the CloudXNet \ref{fig:sixthb} but it reaches lower values.}
\label{fig:b_loss__}
\end{figure}

\subsubsection{SPARCS-based training:}

The same strategy has been applied while changing the training dataset to evaluate more accurately the models. This workflow included the training of the seven deep learning models on SPARCS datasets. Table \ref{results_sparcs} summarizes the evaluation results on the AUC, Dice, IoU and coverage similarity metrics.

\textbf{SPARCS-to-SPARCS evaluation}: When assessing the algorithms on their original training data, the following have been observed:
(1) DeepLabV3+ performs better with the highest values in every metric (AUC of 0.9341, Dice coefficient of 0.8986, IoU of 0.8158, and coverage similarity of 0.9947), demonstrating its efficiency to learn and replicate the patterns from the training dataset. This highlights DeepLabV3+ potential to model complex relationships between inputs and target classes. (2) Closely followed by RS-Net, this model ranks second across most indicators, showing a very Coverage Similarity (0.9965). (3) U-Net++ outperforms both U-Net and Cloud-Net in AUC, Dice and IoU, demonstrating improved generalizability via a nested architecture design. (4) YOLOv8 trails significantly behind other methods with the lowest metric values. 

\textbf{SPARCS-to-Biome evaluation}: Upon evaluating the performances of the algorithms on data acquired from difference sources to test its generalization capacity, in this case test images from Biome dataset, U-Net++ emerges as the top-performing model with the highest AUC, Dice, and IoU. This indicates that U-Net++ generalizes well across different datasets, outperforming other models in accurately segmenting cloud structures. Although U-Net and Cloud-Net experiences slight degradations in certain measures like IoU (0.8144), it still preserves high AUC (0.8880) and Dice coefficients (0.8977). Conversely, RS-Net undergoes substantial reductions across nearly all metrics, dropping to third place in AUC and fifth place in IoU.

\textbf{SPARCS-to-95 Clouds evaluation}:  U-Net++ exhibits strong generalization, achieving the highest scores in Dice (0.6842), IoU (0.5200) and coverage similarity (0.9681). This suggests its adaptability to diverse cloud types and environmental conditions.
As for YOLO, it does not seem to be competing with the other models for remote sensing segmentation as exhibits some very low values. It's noteworthy that YOLOv8 reaches an asymptote very quickly (around 50 epochs) but with the highest loss value compared to the other models.

\begin{table}[H]
\centering
\caption{Evaluation results for the SPARCS trained models.}
\label{results_sparcs}
    \begin{adjustbox}{width=250 pt}
        \begin{threeparttable}[b]
            \small 
            \begin{tabular}{lcccccccc}
                \hline
                Model & AUC & Dice & IoU & Coverage similarity \\ \hline
                \multicolumn{9}{l}{\textbf{Trained on SPARCS ground truth and tested on SPARCS}} \\
                U-Net & 0.9027 & 0.7399 & 0.5871 & 0.9334 \\
                RS-Net & 0.9340 & 0.8813 & 0.7879 & \textbf{0.9965} \\
                U-Net++ & 0.8070 & 0.5670 & 0.3956 & 0.8772 \\
                DeepLabV3+ & \textbf{0.9341} & \textbf{0.8986} & \textbf{0.8158} & 0.9947 \\
                Cloud-Net & 0.9272 & 0.8629 & 0.7589 & 0.9926 \\
                CloudXNet & 0.8910 & 0.7715 & 0.6281 & 0.9715 \\
                YOLOv8 & 0.6373 & 0.4262 & 0.2708 & 0.8971 \\ \hline
                \multicolumn{9}{l}{\textbf{Trained on SPARCS ground truth and tested on Biome}} \\
                U-Net & 0.8880 & 0.8977 & 0.8144 & \textbf{0.9961} \\
                RS-Net & 0.8679 & 0.8593 & 0.7533 & 0.9026 \\
                U-Net++ & \textbf{0.9241} & \textbf{0.9320} & \textbf{0.8726} & 0.994 \\
                DeepLabV3+ & 0.7601 & 0.6996 & 0.5380 & 0.7701 \\
                Cloud-Net & 0.8464 & 0.8338 & 0.7150 & 0.8696 \\
                CloudXNet & 0.7830 & 0.7464 & 0.5955 & 0.8212 \\
                YOLOv8 & 0.5166 & 0.0769 & 0.04 & 0.4788 \\ \hline
                \multicolumn{9}{l}{\textbf{Trained on SPARCS ground truth and tested on 95 Clouds}} \\
                U-Net & 0.6783 & 0.6295 & 0.4593 & 0.9019 \\
                RS-Net & 0.7428 & 0.6751 & 0.5096 & 0.8101 \\
                U-Net++ & 0.7074 & \textbf{0.6842} & \textbf{0.5200} & \textbf{0.9681} \\
                DeepLabV3+ & 0.6809 & 0.5430 & 0.3727 & 0.7149 \\
                Cloud-Net & \textbf{0.7450} & 0.6729 & 0.5070 & 0.7959 \\
                CloudXNet & 0.7094 & 0.6235 & 0.4530 & 0.7917 \\
                YOLOv8 & 0.5251 & 0.1017 & 0.0536 & 0.5543 \\
                \hline
            \end{tabular}
        \end{threeparttable}
    \end{adjustbox}
\end{table}

Similar to the previous training, Fig. \ref{fig:s_loss} shows the loss curves of this second set of models. While YOLO converges very quickly, it has very high values as the convergence asymptote is attained at approximately 1.3 for training and 1.72 for testing. On the other hand, DeepLabv3 demonstrates superior stability among these models followed by Cloud-Net and RS-Net. The models were trained up to 300 epochs because despite having fewer images and information the models seemed to be learning more and for longer from this set.

The obtained results in this second testing phase are highly similar to the previous training where DeepLabV3+ seems to be performing generally good regardless of the training or testing dataset. Followed by U-Net++ and RS-Net, two architectures delivering close results in terms of Dice, IoU and coverage similarity. Cloud-Net and CloudXNet performances have been limited but still delivering decent estimation of the segmentation mask. Finally, YOLOv8 has not shown a good estimation as expected because YOLOv8 performance is unrivaled when it comes to real time operation but in terms of accuracy the other models are better.

\begin{figure}[H]
    \centering
\begin{subfigure}{8cm}
    \includegraphics[width=\textwidth]{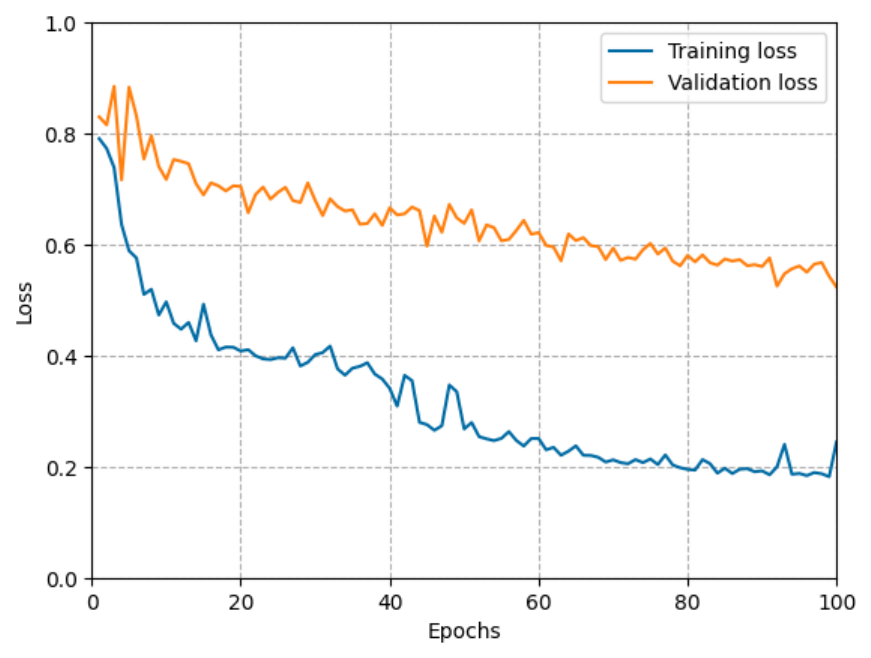}
    \caption{U-Net.}
    \label{fig:first}
\end{subfigure}
\hfill
\begin{subfigure}{8cm}
    \centering
    \includegraphics[width=\textwidth]{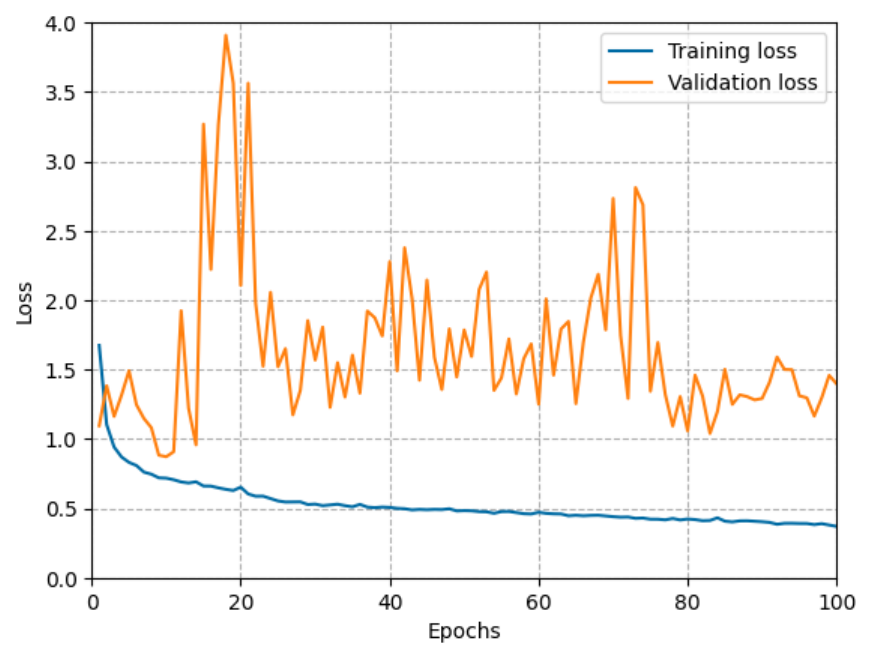}
    \caption{U-net++.}
    \label{fig:second}
\end{subfigure}
\caption{The U-Net \ref{fig:first} confirms his great performances with a small amount of data and the U-Ne++ \ref{fig:second} has very bad stability and high convergence values.}
\label{fig:figures}
\end{figure}
\begin{figure}[H]
\ContinuedFloat
\begin{subfigure}{8cm}
    \centering
    \includegraphics[width=\textwidth]{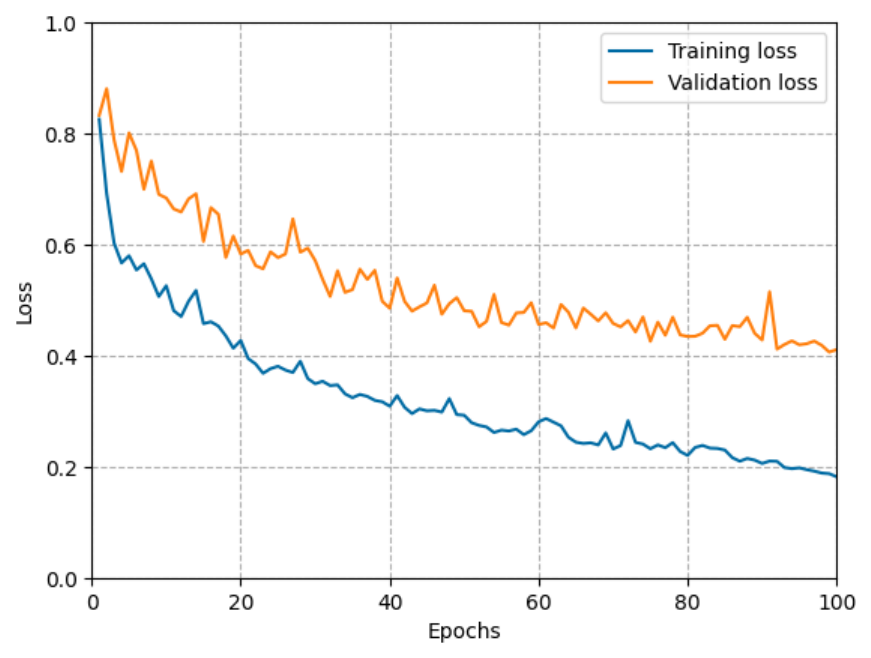}
    \caption{RS-Net.}
    \label{fig:third}
\end{subfigure}
\hfill
\begin{subfigure}{8cm}
    \centering
    \includegraphics[width=\textwidth]{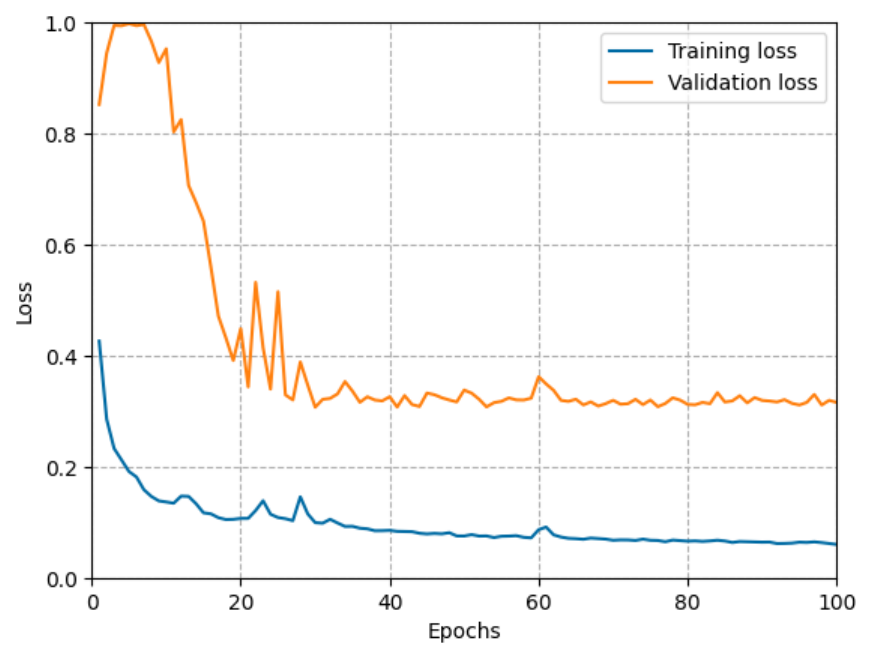}
    \caption{DeepLab V3+.}
    \label{fig:fourth}
\end{subfigure}    
\caption{Both models in \ref{fig:third} and \ref{fig:fourth} perform well, with DeepLabV3+ having slightly lower final values.}
\label{fig:figures}
\end{figure}
\begin{figure}[H]
\ContinuedFloat
\centering
\begin{subfigure}{8cm}
    \centering
    \includegraphics[width=\textwidth]{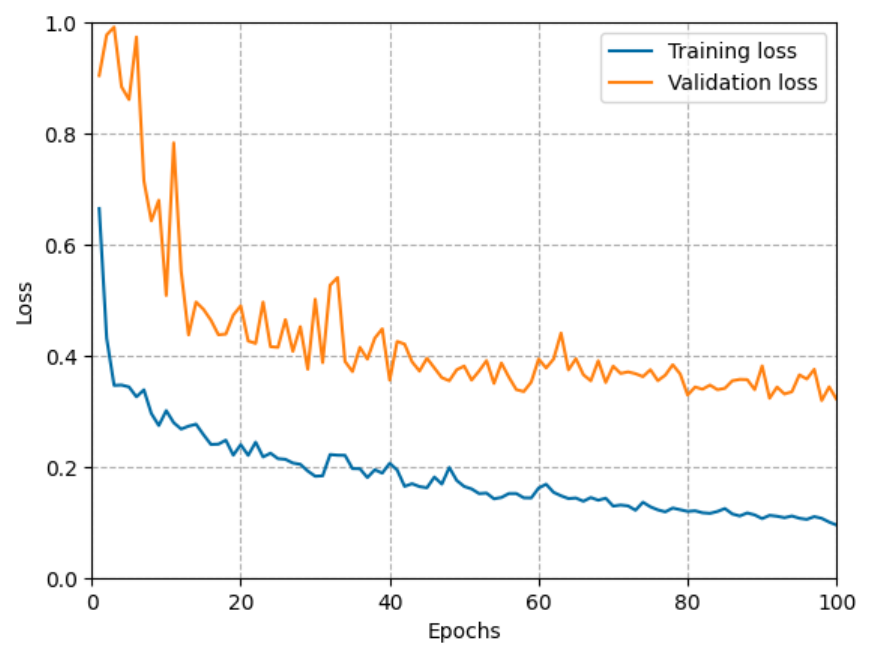}
    \caption{CloudNet.}
    \label{fig:fifth}
\end{subfigure}
\hfill
\begin{subfigure}{8cm}
    \centering
    \includegraphics[width=\textwidth]{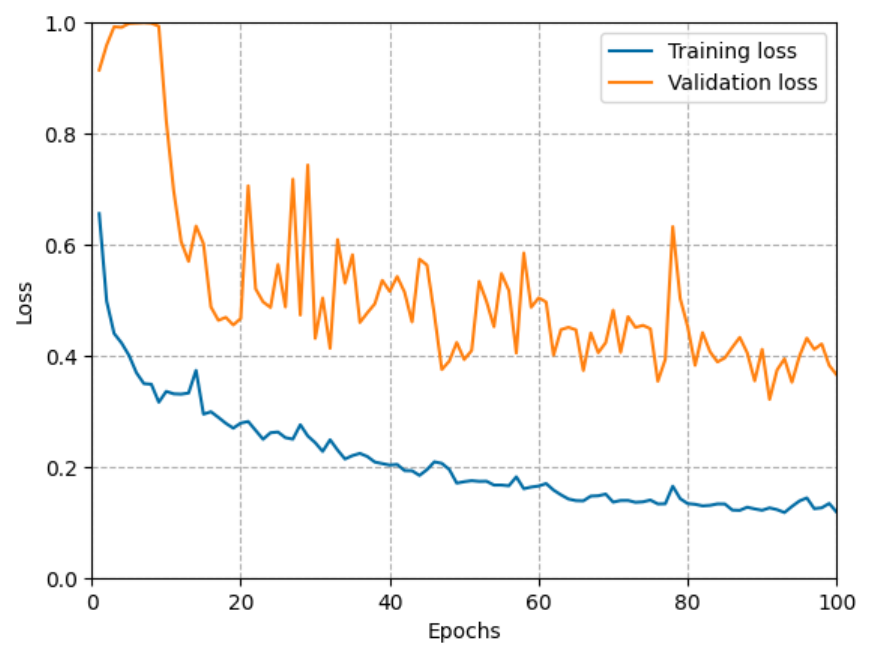}
    \caption{Cloud X Net.}
    \label{fig:sixth}
\end{subfigure}
\caption{The CloudNet model \ref{fig:fifth} has better stability and values than the CloudXNet \ref{fig:sixth}.}
\end{figure}
\begin{figure}[h]
\centering
\ContinuedFloat
\begin{subfigure}{8cm}
    \centering
    \includegraphics[width=\textwidth]{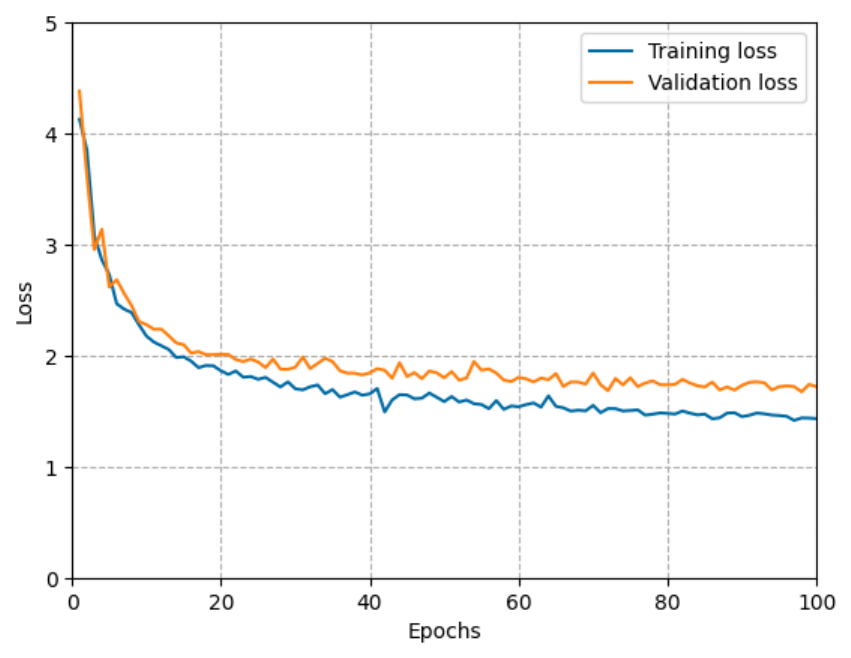}
    \caption{YOLO V8.}
    \label{fig:seventh}
\end{subfigure}
\caption{Similarly to before YOLO \ref{fig:seventh} is fast to converge but has high final values.}
\label{fig:s_loss}
\end{figure}

\subsection{Discussion}

\begin{figure}[H]
    \begin{center}
        \includegraphics[width=0.9\textwidth]{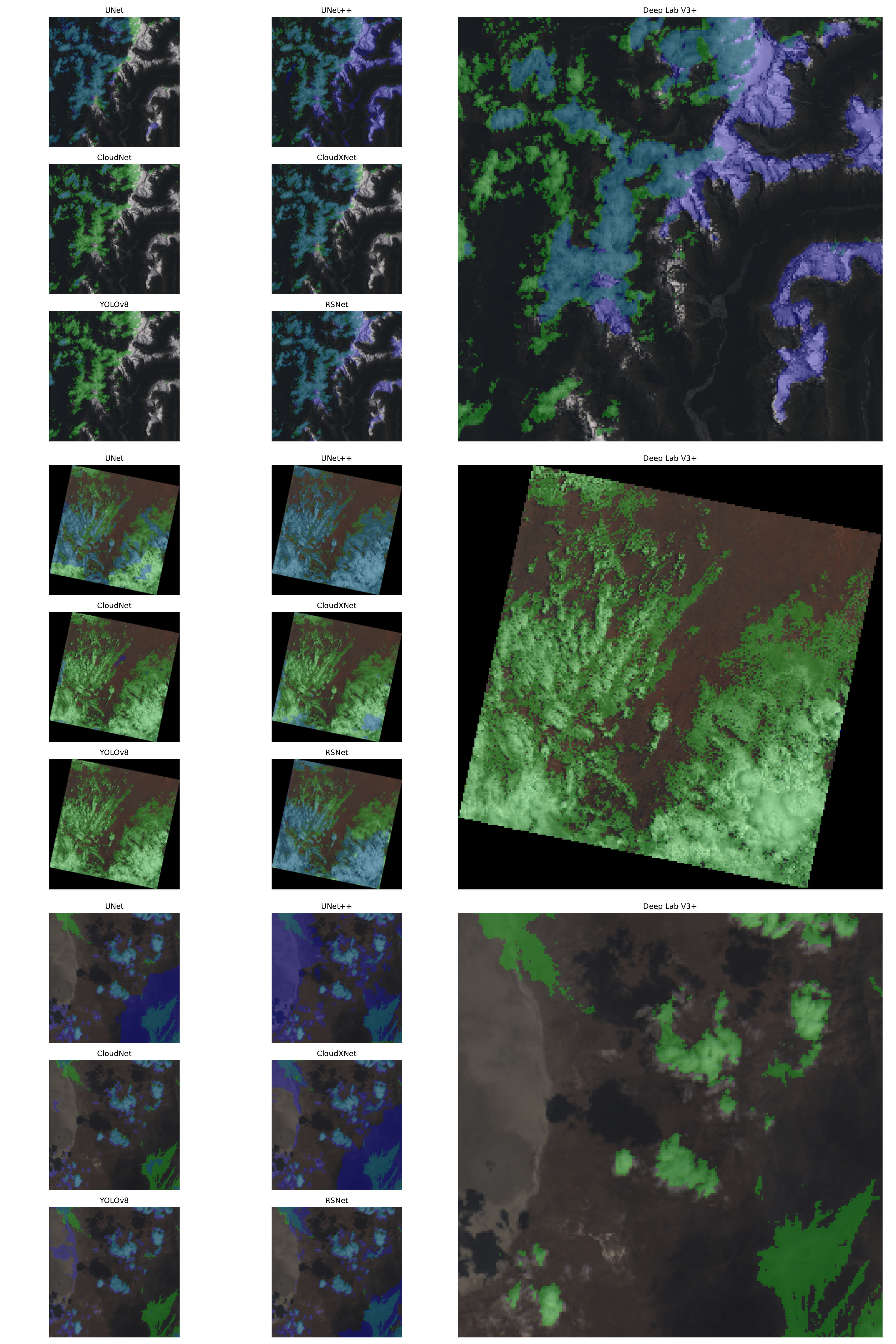}
    \end{center}
    \caption{Prediction comparison, taken from scenes LC80760912014013LGN00, LC81750512013208LGN00, LC08L1TP045020201607202017022101T1 (top to bottom). Models trained with \textbf{Biome} ground truth. Prediction in blue, ground truth in green. When the two colours overlap, the prediction is correct.}
    \label{fig:b_im}
\end{figure}

\begin{figure}[H]
    \begin{center}
        \includegraphics[width=0.9\textwidth]{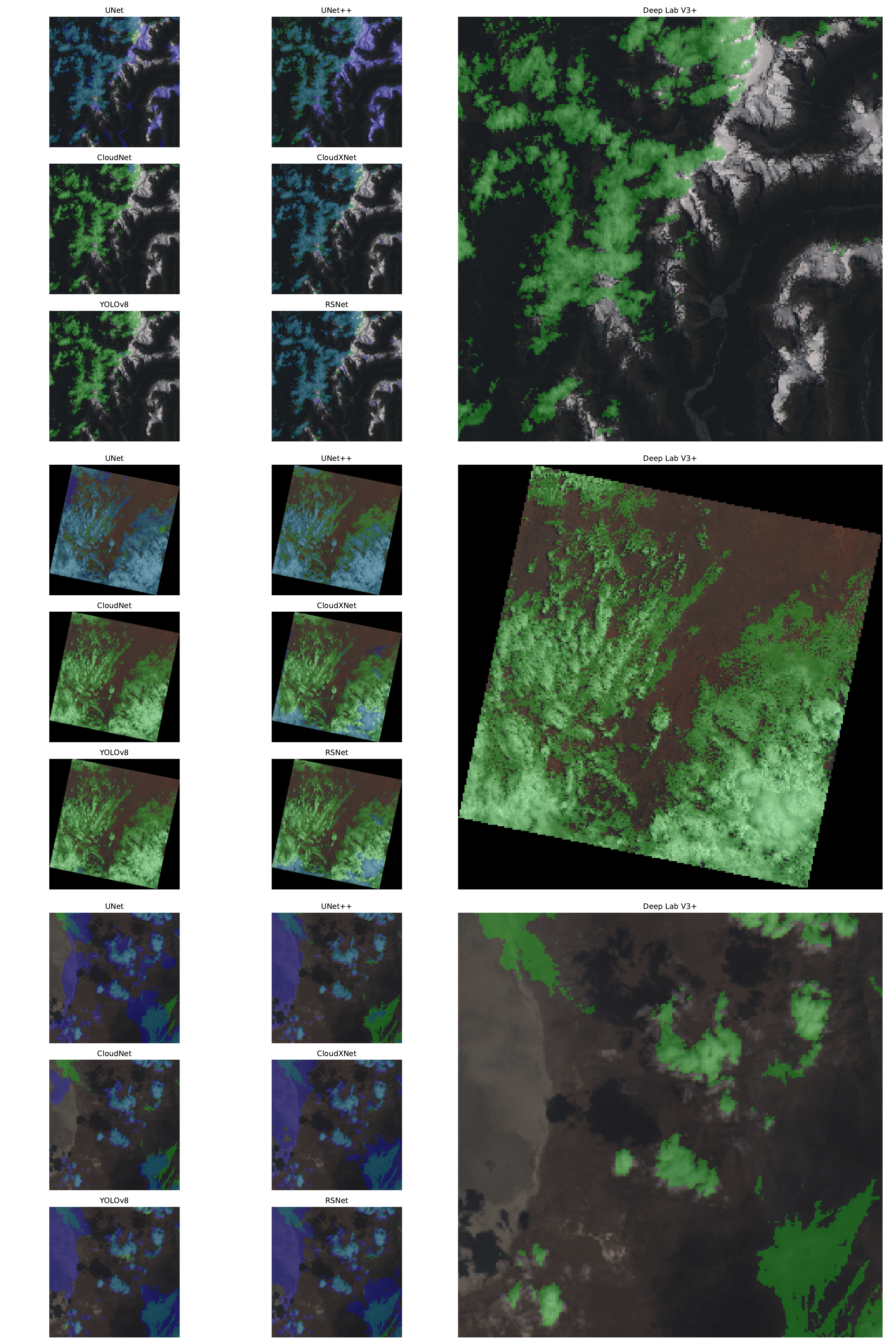}
    \end{center}
    \caption{Prediction comparison, taken from scenes LC80760912014013LGN00, LC81750512013208LGN00, LC08L1TP045020201607202017022101T1 (top to bottom). Models trained with \textbf{SPARCS} ground truth. Prediction in blue, ground truth in green. When the two colours overlap, the prediction is correct.}
    \label{fig:sp_im}
\end{figure}

The performance trends observed in Table \ref{results_sparcs} are consistent with those in Table \ref{results_biome}, highlighting DeepLabV3+ and RS-Net as consistently strong performers, while U-Net and U-Net++ exhibit relatively lower performance compared to other models. The specific dataset used for training appears to influence model performance, emphasizing the importance of dataset compatibility in deep learning applications. In addition to the numerical measures, visualisations provide a tactile peek into the algorithms' real-world performance. Visual representation of the segmentation model applied on some extreme cases test image is represented in Fig.\ref{fig:b_im} and Fig.\ref{fig:sp_im}, where it shows that the performances of DeepLabV3+ and RS-Net isn't always outperforming the other models.

It is noteworthy to mention, from this comparison investigation, the impact of the exact dataset used during training on the subsequent performance of the segmentation models. The intricacies embedded in each dataset shape the capabilities and limitations of the deployed algorithms. This result could be observed in the first scene \textit{LC80760912014013LGN00} as the \textbf{Biome}-based training has led to an estimation of the clouds while no clouds has been detected on the same scene when tested on \textbf{SPARCS}-based trained models. Which underlines the critical importance of dataset compatibility in the complex terrain of deep learning applications, highlighting the need for a nuanced understanding of dataset features and their possible impact on model outcomes.

\section{Conclusion}

This benchmark study thoroughly evaluated the performance of seven segmentation algorithms applied to remote sensing, with a particular focus on cloud segmentation using the Biome and SPARCS datasets. Evaluating various metrics, including AUC, Dice coefficient, IoU, and coverage similarity, revealed differentiated strengths and weaknesses of each algorithm in different scenarios. Notably, RS-Net and DeepLabV3+ consistently showed robust performance across different datasets, demonstrating their versatility in cloud segmentation tasks. 
The results highlight the importance of careful algorithm selection tailored to specific dataset characteristics, with RS-Net emerging as a reliable choice for cross-dataset evaluations. The results provide valuable insights for remote sensing practitioners and researchers and highlight the need for algorithmic adaptability to ensure optimal performance in different cloud segmentation scenarios. Future research directions could include further fine-tuning of algorithms and exploring new approaches to improve segmentation accuracy in specific remote sensing applications.

In conclusion, although these algorithms have demonstrated commendable performance, there are several promising avenues for further enhancements. One such avenue involves leveraging ensemble methods, which amalgamate the outputs of multiple models to enhance predictive performance. Additionally, exploring novel architectures, whether based on the U-Net framework or entirely new designs, offers opportunities for improvement. Ongoing research on hyperparameter optimization, including the refinement of network architectures, contributes to the continuous evolution of the field.

Furthermore, integrating temporal patterns into the model holds significance, especially in the context of remote sensing where static buildings, distinct vegetation cycles, and unpredictable cloud behaviours are prevalent. Despite the current computational challenges associated with multi-temporal semantic segmentation methods based on deep learning, recent research indicates promising outcomes. Anticipated future developments are expected to produce methods seamlessly integrating spectral, spatial, and temporal dimensions, thereby advancing the field.

\bibliographystyle{unsrt}  
\bibliography{references}

\begin{thebibliography}{10}

\bibitem{foga_cloud_2017}
Steve Foga, Pat~L. Scaramuzza, Song Guo, Zhe Zhu, Ronald~D. Dilley, Tim Beckmann, Gail~L. Schmidt, John~L. Dwyer, M.~Joseph~Hughes, and Brady Laue.
\newblock Cloud detection algorithm comparison and validation for operational {Landsat} data products.
\newblock {\em Remote Sensing of Environment}, 194:379--390, June 2017.

\bibitem{mohajerani_cloud_2018}
S.~Mohajerani, T.~A. Krammer, and P.~Saeedi.
\newblock "{A} {Cloud} {Detection} {Algorithm} for {Remote} {Sensing} {Images} {Using} {Fully} {Convolutional} {Neural} {Networks}".
\newblock In {\em {IEEE} 20th {International} {Workshop} on {Multimedia} {Signal} {Processing} ({MMSP})}, pages 1--5, August 2018.
\newblock ISSN: 2473-3628.

\bibitem{mlhub_sentinel_2}
De~Felice L, Pekel JF, Ugolotti R, and Kempeneers P.
\newblock Sentinel-2 global-scale open access labelled dataset, 2023.

\bibitem{jeppesen_cloud_2019}
Jacob~Høxbroe Jeppesen, Rune~Hylsberg Jacobsen, Fadil Inceoglu, and Thomas~Skjødeberg Toftegaard.
\newblock A cloud detection algorithm for satellite imagery based on deep learning.
\newblock {\em Remote Sensing of Environment}, 229:247--259, August 2019.

\bibitem{ronneberger_u-net_2015}
Olaf Ronneberger, Philipp Fischer, and Thomas Brox.
\newblock U-{Net}: {Convolutional} {Networks} for {Biomedical} {Image} {Segmentation}, May 2015.
\newblock arXiv:1505.04597 [cs].

\bibitem{DBLP:journals/corr/abs-1807-10165}
Zongwei Zhou, Md~Mahfuzur~Rahman Siddiquee, Nima Tajbakhsh, and Jianming Liang.
\newblock Unet++: {A} nested u-net architecture for medical image segmentation.
\newblock {\em CoRR}, abs/1807.10165, 2018.

\bibitem{ZHOU2021115625}
Junsheng Zhou, Yiwen Lu, Siyi Tao, Xuan Cheng, and Chenxi Huang.
\newblock E-res u-net: An improved u-net model for segmentation of muscle images.
\newblock {\em Expert Systems with Applications}, 185:115625, 2021.

\bibitem{AKTER2024122347}
Atika Akter, Nazeela Nosheen, Sabbir Ahmed, Mariom Hossain, Mohammad~Abu Yousuf, Mohammad Ali~Abdullah Almoyad, Khondokar~Fida Hasan, and Mohammad~Ali Moni.
\newblock Robust clinical applicable cnn and u-net based algorithm for mri classification and segmentation for brain tumor.
\newblock {\em Expert Systems with Applications}, 238:122347, 2024.

\bibitem{boureau2010theoretical}
Y-Lan Boureau, Jean Ponce, and Yann LeCun.
\newblock A theoretical analysis of feature pooling in visual recognition.
\newblock In {\em Proceedings of the 27th international conference on machine learning (ICML-10)}, pages 111--118, 2010.

\bibitem{sym13122246}
Tomasz Hachaj, Anna Stolińska, Magdalena Andrzejewska, and Piotr Czerski.
\newblock Deep convolutional symmetric encoder\&mdash;decoder neural networks to predict students\&rsquo; visual attention.
\newblock {\em Symmetry}, 13(12), 2021.

\bibitem{mao2016image}
Xiaojiao Mao, Chunhua Shen, and Yu-Bin Yang.
\newblock Image restoration using very deep convolutional encoder-decoder networks with symmetric skip connections.
\newblock {\em Advances in neural information processing systems}, 29, 2016.

\bibitem{zhou_unet_2018}
Zongwei Zhou, Md~Mahfuzur~Rahman Siddiquee, Nima Tajbakhsh, and Jianming Liang.
\newblock {UNet}++: {A} {Nested} {U}-{Net} {Architecture} for {Medical} {Image} {Segmentation}, July 2018.
\newblock arXiv:1807.10165 [cs, eess, stat].

\bibitem{hariharan2016object}
Bharath Hariharan, Pablo Arbelaez, Ross Girshick, and Jitendra Malik.
\newblock Object instance segmentation and fine-grained localization using hypercolumns.
\newblock {\em IEEE transactions on pattern analysis and machine intelligence}, 39(4):627--639, 2016.

\bibitem{Huang_2017_CVPR}
Gao Huang, Zhuang Liu, Laurens van~der Maaten, and Kilian~Q. Weinberger.
\newblock Densely connected convolutional networks.
\newblock In {\em Proceedings of the IEEE Conference on Computer Vision and Pattern Recognition (CVPR)}, July 2017.

\bibitem{kaur_buttar_semantic_2022}
Preetpal Kaur~Buttar and Manoj~Kumar Sachan.
\newblock Semantic segmentation of clouds in satellite images based on {U}-{Net}++ architecture and attention mechanism.
\newblock {\em Expert Systems with Applications}, 209:118380, December 2022.

\bibitem{zhu_object-based_2012}
Zhe Zhu and Curtis~E. Woodcock.
\newblock Object-based cloud and cloud shadow detection in {Landsat} imagery.
\newblock {\em Remote Sensing of Environment}, 118:83--94, March 2012.

\bibitem{DBLP:journals/corr/abs-2009-07485}
Hossein Gholamalinezhad and Hossein Khosravi.
\newblock Pooling methods in deep neural networks, a review.
\newblock {\em CoRR}, abs/2009.07485, 2020.

\bibitem{DBLP:journals/corr/DrozdzalVCKP16}
Michal Drozdzal, Eugene Vorontsov, Gabriel Chartrand, Samuel Kadoury, and Chris Pal.
\newblock The importance of skip connections in biomedical image segmentation.
\newblock {\em CoRR}, abs/1608.04117, 2016.

\bibitem{reddi_convergence_2019}
Sashank~J. Reddi, Satyen Kale, and Sanjiv Kumar.
\newblock On the {Convergence} of {Adam} and {Beyond}.
\newblock April 2019.
\newblock arXiv:1904.09237 [cs, math, stat].

\bibitem{kingma_adam_2017}
Diederik~P. Kingma and Jimmy Ba.
\newblock Adam: {A} {Method} for {Stochastic} {Optimization}.
\newblock January 2017.
\newblock arXiv:1412.6980 [cs].

\bibitem{reis_real-time_2023}
Dillon Reis, Jordan Kupec, Jacqueline Hong, and Ahmad Daoudi.
\newblock Real-{Time} {Flying} {Object} {Detection} with {YOLOv8}, May 2023.
\newblock arXiv:2305.09972 [cs].

\bibitem{WANG2022116793}
Yi~Wang, Syed Muhammad~Arsalan Bashir, Mahrukh Khan, Qudrat Ullah, Rui Wang, Yilin Song, Zhe Guo, and Yilong Niu.
\newblock Remote sensing image super-resolution and object detection: Benchmark and state of the art.
\newblock {\em Expert Systems with Applications}, 197:116793, 2022.

\bibitem{ZHOU2024122256}
Yan Zhou.
\newblock A yolo-nl object detector for real-time detection.
\newblock {\em Expert Systems with Applications}, 238:122256, 2024.

\bibitem{photon_revolutionizing_2023}
Chang~Ho Kang and Sun~Young Kim.
\newblock Real-time object detection and segmentation technology: an analysis of the yolo algorithm.
\newblock {\em JMST Advances}, 5(2):69--76, 9 2023.
\newblock SN 2524-7913.

\bibitem{chen2018encoderdecoder}
Liang-Chieh Chen, Yukun Zhu, George Papandreou, Florian Schroff, and Hartwig Adam.
\newblock Encoder-decoder with atrous separable convolution for semantic image segmentation, 2018.

\bibitem{chen2017rethinking}
Liang-Chieh Chen, George Papandreou, Florian Schroff, and Hartwig Adam.
\newblock Rethinking atrous convolution for semantic image segmentation, 2017.

\bibitem{chen2017deeplab}
Liang-Chieh Chen, George Papandreou, Iasonas Kokkinos, Kevin Murphy, and Alan~L. Yuille.
\newblock Deeplab: Semantic image segmentation with deep convolutional nets, atrous convolution, and fully connected crfs, 2017.

\bibitem{chen2016semantic}
Liang-Chieh Chen, George Papandreou, Iasonas Kokkinos, Kevin Murphy, and Alan~L. Yuille.
\newblock Semantic image segmentation with deep convolutional nets and fully connected crfs, 2016.

\bibitem{lin2017feature}
Tsung-Yi Lin, Piotr Dollár, Ross Girshick, Kaiming He, Bharath Hariharan, and Serge Belongie.
\newblock Feature pyramid networks for object detection, 2017.

\bibitem{fu2017dssd}
Cheng-Yang Fu, Wei Liu, Ananth Ranga, Ambrish Tyagi, and Alexander~C. Berg.
\newblock Dssd : Deconvolutional single shot detector, 2017.

\bibitem{shrivastava2017skip}
Abhinav Shrivastava, Rahul Sukthankar, Jitendra Malik, and Abhinav Gupta.
\newblock Beyond skip connections: Top-down modulation for object detection, 2017.

\bibitem{pinheiro2016learning}
Pedro~O. Pinheiro, Tsung-Yi Lin, Ronan Collobert, and Piotr Dollàr.
\newblock Learning to refine object segments, 2016.

\bibitem{jiang2018rednet}
Jindong Jiang, Lunan Zheng, Fei Luo, and Zhijun Zhang.
\newblock Rednet: Residual encoder-decoder network for indoor rgb-d semantic segmentation, 2018.

\bibitem{mpdis21217191}
Samee~Ullah Khan, Ijaz~Ul Haq, Zulfiqar~Ahmad Khan, Noman Khan, Mi~Young Lee, and Sung~Wook Baik.
\newblock Atrous convolutions and residual gru based architecture for matching power demand with supply.
\newblock {\em Sensors}, 21(21), 2021.

\bibitem{howard2017mobilenets}
Andrew~G. Howard, Menglong Zhu, Bo~Chen, Dmitry Kalenichenko, Weijun Wang, Tobias Weyand, Marco Andreetto, and Hartwig Adam.
\newblock Mobilenets: Efficient convolutional neural networks for mobile vision applications, 2017.

\bibitem{mohajerani_cloud-net_2019}
Sorour Mohajerani and Parvaneh Saeedi.
\newblock Cloud-{Net}: {An} end-to-end {Cloud} {Detection} {Algorithm} for {Landsat} 8 {Imagery}, January 2019.
\newblock arXiv:1901.10077 [cs].

\bibitem{lu2022dual}
Chen Lu, Min Xia, Ming Qian, and Binyu Chen.
\newblock Dual-branch network for cloud and cloud shadow segmentation.
\newblock {\em IEEE Transactions on Geoscience and Remote Sensing}, 60:1--12, 2022.

\bibitem{nair_rectified_nodate}
Geoffrey E.~Hinton Vinod~Nair.
\newblock Rectified linear units improve restricted boltzmann machines.
\newblock pages 807--814, 2010.

\bibitem{kanu_cloudx-net_2020}
Sumit Kanu, Rohit Khoja, Shyam Lal, B.~S. Raghavendra, and Asha Cs.
\newblock {CloudX}-net: {A} robust encoder-decoder architecture for cloud detection from satellite remote sensing images.
\newblock {\em Remote Sensing Applications: Society and Environment}, 20:100417, November 2020.

\bibitem{Mller2022TowardsAG}
Dominik M{\"u}ller, I{\~n}aki Soto-Rey, and Frank Kramer.
\newblock Towards a guideline for evaluation metrics in medical image segmentation.
\newblock {\em BMC Research Notes}, 15, 2022.

\bibitem{Basu2023}
Anurag Basu, Purnendu Senapati, Monojit Deb, Rishiraj Rai, and Krishnendu~G. Dhal.
\newblock A survey on recent trends in deep learning for nucleus segmentation from histopathology images.
\newblock 2023:1--46, Mar 2023.

\bibitem{BRADLEY19971145}
Andrew~P. Bradley.
\newblock The use of the area under the roc curve in the evaluation of machine learning algorithms.
\newblock {\em Pattern Recognition}, 30(7):1145--1159, 1997.

\bibitem{ZOU2004178}
Kelly~H. Zou, Simon~K. Warfield, Aditya Bharatha, Clare~M.C. Tempany, Michael~R. Kaus, Steven~J. Haker, William~M. Wells, Ferenc~A. Jolesz, and Ron Kikinis.
\newblock Statistical validation of image segmentation quality based on a spatial overlap index1: scientific reports.
\newblock {\em Academic Radiology}, 11(2):178--189, 2004.

\bibitem{Rahmaniou2016}
Md~Atiqur Rahman and Yang Wang.
\newblock Optimizing intersection-over-union in deep neural networks for image segmentation.
\newblock In {\em Advances in Visual Computing}, pages 234--244, Cham, 2016. Springer International Publishing.

\end{thebibliography}

\end{document}